\RequirePackage{fix-cm}
\documentclass[smallextended]{for_arxiv}    
\smartqed  
\usepackage{graphicx}
\usepackage{array,amssymb,amsmath,amsfonts}
\usepackage[authoryear]{natbib}

\usepackage{subfigure}
\usepackage{multirow}
\usepackage{math}
\usepackage{tikz}
\usetikzlibrary{arrows,automata,shapes}

\usepackage{bm,color}

\newcommand{\black}[1]{\textcolor{black}{#1}}

\graphicspath{{../MMST/figures/}{../MMST/IMAGES/}}
\begin{document}

\pagestyle{plain}

\def\norm#1{\left\|#1\right\|}             \def\stdnorm#1{\|#1\|}

\title{High-Dimensional Regression with Gaussian Mixtures and Partially-Latent Response Variables
}

\titlerunning{High-Dimensional Regression with Mixtures and Partially-Latent Variables}        

\author{Antoine Deleforge       \and
        Florence Forbes         \and 
        Radu Horaud
}


\institute{A. Deleforge, F. Forbes, R. Horaud \at
           INRIA Grenoble Rh\^one-Alpes\\ 655 avenue de l'Europe\\ 38330 Montbonnot Saint-Martin, France \\
           Tel.: +33 4 76 61 52 08\\
           \email{firstname.lastname@inria.fr}
}

\maketitle

\begin{abstract}
In this work we address the problem of approximating high-di\-men\-sio\-nal data with a low-dimensional representation. We make the following contributions. We propose an inverse regression method which exchanges the roles of input and response, such that the low-dimensional variable becomes the regressor, and which is tractable. We introduce a mixture of locally-linear probabilistic mapping model that starts with estimating the parameters of inverse regression,  and follows
 with inferring closed-form solutions for the forward parameters of the high-dimensional regression problem of interest. Moreover, we introduce a partially-latent paradigm, such that the vector-valued response variable is composed of both observed and latent entries, thus being able to deal with data contaminated by experimental artifacts that cannot be explained with noise models. The proposed probabilistic formulation could be viewed as a
latent-variable augmentation of regression. We devise 
expectation-maximization (EM) procedures based on a data
augmentation strategy which facilitates the
maximum-likelihood search over the model parameters. We
propose two augmentation schemes and we describe in detail the associated
EM inference procedures that may well be viewed as generalizations of a number of
EM regression, dimension reduction, and factor analysis algorithms. The proposed framework is validated with both synthetic and real data. We provide experimental evidence that our method outperforms several existing regression techniques.
\end{abstract}
\newpage
\section{Introduction}
\label{sec:intro}
The task of regression consists of learning a mapping from an input variable onto a response variable, such that the response of a test point 
could be easily and robustly computed. While this problem has been extensively studied, situations where the input variable is of high dimension, and where the response variable may not be fully observed, still challenge the current state of the art. 
It is well known that high-dimensional to low-dimensional (or high-to-low) regression is problematic, and usually performed in two separated steps: dimensionality reduction followed by regression.
\color{black}
In this paper we propose a novel formulation whose originality is twofold: (i)~it overcomes the difficulties of high-to-low regression by exchanging the roles of the input and response variables, and (ii)~it incorporates a partially-latent (or partially-observed) response variable model that captures unobserved data.

To bypass the problems associated with high-to-low regression, the roles of the input and response variables are exchanged, such that \textit{the low-dimensional variable becomes the regressor}. 
We start by estimating the parameters of a low-to-high regression model, or inverse regression \citep{Li91}, \textcolor{black}{from which we derive} the \textit{forward parameters} \textcolor{black}{characterizing} the high-to-low regression problem of interest. We show that, by using mixture models, this inverse-then-forward strategy becomes tractable.
Moreover, we allow the low-dimensional variable to be only partially observed (or equivalently, partially latent), \textit{i.e.}, \textit{the vector-valued low-dimensional variable is composed of both observed entries and latent entries}. This is particularly relevant for a variety of applications where the data are too complex to be totally observed.
\color{black}
\color{black}
Starting from standard mixture of linear regressions,
we propose a novel mixture of locally-linear regression model that unifies regression and dimensionality reduction into a common framework. 
\color{black}
The 
probabilistic formulation that we derive may be seen as a
latent-variable augmentation of regression. We devise an
associated expectation-maximization procedure based on a data
augmentation strategy, thus facilitating the subsequent
maximum-likelihood search over the model parameters. We
investigate two augmentation schemes and, in practice, we propose two
EM algorithms that can be viewed as generalizations of a number of
EM algorithms either for regression or for dimension reduction. The
proposed method is particularly interesting for solving
high-to-low regression problems in the
presence of training data corrupted by irrelevant information. It has the potential of dealing with many
applications, where the response variable can only be partially observed,
either because it cannot be measured with appropriate sensors, or
because it cannot be easily annotated. In other terms, the proposed method allows
a form of \textit{slack} in the response vector by adding a
few latent entries to the vector's observed entries.

The remainder of this paper is organized as follows. Related work, background, and contributions are described in Section~\ref{sec:background}. 
Section~\ref{sec:gllim} describes in detail the proposed \textit{Gaussian locally-linear mapping} (GLLiM) model, which solves for \textit{inverse regression}, and derives the formulae for \textit{forward regression}.
Next, Section~\ref{sec:plvm} shows how to incorporate a partially-latent variable into GLLiM and discusses the link with 
a number of existing regression and dimensionality reduction techniques. Section~\ref{sec:algo} describes the
proposed expectation-maximization framework for estimating the parameters of
the model. Section~\ref{sec:results} describes the experimental validation of our method and compares it with a number of state-of-the-art regression techniques using
synthetic data, a dataset of 3D
faces, and a dataset of hyper-spectral images of Mars surface. Finally,
Section~\ref{sec:conclusion} concludes with a discussion and future directions
of research.
\black{In addition, a companion document referred to as ``Supplementary Material" provides details omitted in the manuscript. This document as well as Matlab code for the proposed algorithms and illustrative examples are available online\footnote{\black{https://team.inria.fr/perception/gllim\_toolbox/.}}.}

\section{Related Work, Background, and Contributions}
\label{sec:background}
\subsection{Dealing with High-Dimensional Data}
\label{subsec:intro_high_dim}
Estimating a function defined over a space of high dimension, say $D$, is generally hard because standard regression methods 
have to estimate a large number of parameters, typically of the order of $D^2$. For this reason, existing methods proceed in two steps: dimension
reduction followed by regression. This sequential way of doing presents the risk to map the
input onto an intermediate low-dimensional space that
does not necessarily contain the information needed to correctly predict the
output. To prevent this problem, a number of methods perform
the dimension reduction step by taking the output variable
into account. The concept of {\it sufficient
reduction} \citep{Cook07} was specifically introduced for solving regression problems of this type. The process of replacing the input with a
lower-dimensional representation is called {\it sufficient dimension
reduction} which retains all relevant information
about the output. Methods falling into this category are partial
least-squares (PLS) \citep{rosipal2006overview}, sliced inverse
regression (SIR) \citep{Li91}, kernel SIR \citep{wu2008kernel}, and principal component based methods \citep{Cook07,AdragniCook2009}. SIR
methods are not designed specifically for prediction and do not
provide a specific predictive method.
Once a dimension reduction
has been determined, any standard method can then be used to perform
predictions, which are likely to be sub-optimal since they are not
necessarily consistent with the reduction model. Regarding PLS,
its superior performance over standard principal component regression
is subject to the relationship between the covariances of input and output variables, and the eigen-structure of the covariance of
the input variables \citep{NaikTsai2000}. The principal component
methods proposed in \citep{Cook07,AdragniCook2009} are based on a
semi-parametric model of
the input given the output and can be
used without specifying a model for the joint distribution of input and output variables. By achieving regression in two steps, these approaches cannot be conveniently expressed in terms of a single optimization problem.

\color{black}
We propose a method that bypasses the difficulty of high-to-low regression by considering the problem the other way around, \textit{i.e.}, low-to-high. We denote the low-dimensional data with $\{\xvect_n\}_{n=1}^N\subset\mathbb{R}^L$, the high-dimensional data with $\{\yvect_n\}_{n=1}^N\subset\mathbb{R}^D$ ($D\gg L$), and we assume that these data are realizations of two random variables $\Xvect$ and $\Yvect$ with joint probability distribution $p(\Yvect,\Xvect;\thetavect)$, where $\thetavect$ denotes the model parameters. At training, the low-dimensional variable $\Xvect$ will play the role of the \textit{regressor}, namely $\Yvect$ \textit{is a function of} $\Xvect$ possibly corrupted by noise through $p(\Yvect|\Xvect;\thetavect)$. Hence, $\Yvect$ \textit{is assumed to lie on a low-dimensional manifold embedded in} $\mathbb{R}^D$ and parameterized by $\Xvect$. The low dimension of the regressor $\Xvect$ will imply a relatively small number of parameters to be estimated, \textit{i.e.}, approximately linear in $L(D+L)$, thus facilitating the task of estimating the model parameters. Once $\thetavect$ has been estimated, we show that the computation of the \textit{forward conditional density} $p(\Xvect|\Yvect;\thetavect)$ is tractable, and hence is used to predict the low-dimensional response $\xvect$ of a high-dimensional test point $\yvect$. This \textit{inverse-then-forward} regression strategy, thoroughly detailed in Section \ref{sec:gllim},  justifies the unconventional notations: $\Yvect$ for the high-dimensional \textit{input} and $\Xvect$ for the low-dimensional \textit{response}.
\color{black}

\subsection{Dealing with Non-Linear Data}
\label{subsec:intro_non_lin}
A classical approach to deal with non-linear data is to use kernel methods. 
These methods map the data onto high-dimensional, possibly infinite, feature spaces.
This is achieved by defining a kernel function over the observation space.
Since the kernel function is often not linear, the relations found in this way are accordingly very general. 
Examples of kernel methods for regression are
kernel SIR \citep{wu2008kernel}, the relevance vector machine method
\citep{tipping2001sparse} or its multivariate extension
\citep{thayananthan2006multivariate}.
Among kernel methods, Gaussian process latent variable models (GPLVM) form a widely used family of probabilistic models. GPLVM was originally formulated
as a dimensionality reduction technique \citep{lawrence2005}. It can be viewed as an instance of non-linear probabilistic principal component analysis. GPLVM was then extended to regression \citep{fusi2012joint,WangNeal12}.
One drawback of all kernel methods is that they require a choice for an appropriate kernel function, which is done in an ad-hoc manner and which are highly application- and data-dependent. Moreover, as pointed out in \citep{lawrence2005}, the mappings learned with kernel methods cannot be inverted. 

Another attractive approach for modeling non-linear data is to use a
mixture of \textit{locally linear} models. In the Gaussian case, this boils down to estimating a Gaussian mixture model (GMM) on the joint input-response variable.
We will refer to the corresponding family of mappings as \textit{supervised} Gaussian Locally-Linear Mapping (GLLiM) in the case of regression, \textit{i.e.}, $\Xvect$ is fully observed, and \textit{unsupervised}
GLLiM in the case of dimensionality reduction, \textit{i.e.}, $\Xvect$ is fully unobserved. 
\black{
Supervised GLLiM may be viewed as an affine instance of mixture of experts as formulated in \citep{XuJordanHinton95} or as cluster-weighted modeling (CWM) \citep{gershenfeld1997nonlinear} except that the response variable is multivariate in GLLiM and scalar in CWM. Interestingly, \citep{ingrassia2012local} recently proposed an extension of CWM to Student-$t$ distributions. However, they do not address high-dimensional regression and they do not consider a partially-latent variable model.
It is worth mentioning that \citep{ingrassia2012local} and \citep{DeleforgeHoraud-MLSP12} provide similar geometric interpretations of these mixture models.
}
In Section \ref{sec:plvm} we point out that a number of other regression methods \citep{quandt1978estimating,deVeaux89,XuJordanHinton95,jedidi1996estimating,KainMacon98,QiaoMinematsu09,DeleforgeHoraud-MLSP12} may be viewed as supervised GLLiM methods, while some dimensionality reduction and factor analysis methods \citep{TippingBishop99,tipping1999probabilistic,GhahramaniHinton96,wedel2001factor,bach2005probabilistic,Bishop98,kalaitzis2011residual} may be viewed as unsupervised GLLiM methods.

\subsection{Dealing with Partially-Observed Response Variables}
\label{subsec:intro_plom}
We propose a generalization of unsupervised and supervised GLLiM referred to as \textit{hybrid GLLiM}.
While the high-dimensional variable $\Yvect$ remains fully
observed, the low-dimensional variable $\Xvect$
is a concatenation of \textit{observed entries}, collectively denoted by
$\Tvect$, and \textit{latent entries}, collectively denoted by $\Wvect$, namely $\Xvect=\left[
 \begin{array}{c}
  \Tvect \; ;
  \Wvect
 \end{array}
\right] $, where $[.;.]$ denotes vertical vector concatenation.
The hybrid GLLiM model is particularly interesting for solving
regression problems in the
presence of data corrupted by irrelevant  information for
the problem at hand. It has the potential of being well suited in many
application scenarios, namely whenever the response variable is only partially observed,
because it is neither available, nor observed with appropriate sensors. The idea of the hybrid GLLiM model is to allow
some form of \textit{slack} by adding a
few latent entries to the response variable. 

\subsection{Application Scenarios}

To further motivate the need for such a model, we consider a few examples.
Motion capture methods use regression to infer a map from
high-dimensional visual data onto a small number of human-joint angles involved in a particular motion being trained, {\it e.g.},
\citep{agarwal2004learning,agarwal2006recovering}. Nevertheless,
the input data contain irrelevant information, such as lighting
effects responsible for various artifacts, which aside from the
fact that it is not relevant for the task at hand, is almost
impossible to be properly modeled, quantified or even annotated.
The recovered low-dimensional representation should
account for such phenomena that are unobservable.

In the field of planetology, hyper-spectral imaging
is used to  recover parameters associated with the physical
properties of planet surfaces
{\it e.g.}, \citep{bernard2009retrieval}. To this end,
radiative transfer models have been developed, that link the
chemical composition, the granularity, or the physical state, to the
observed spectrum. They are generally used to simulate huge
collections of spectra in order to perform the inversion
of hyperspectral images \citep{Doute2007}. As the required
computing resources to generate such a database increases
exponentially with the number of parameters, they are generally
restricted to a small number of parameters, {\it e.g.}, abundance
and grain size of the main chemical components. Other parameters, such
as those related to meteorological variability or the incidence angle of the
spectrometer are neither explicitly modeled nor measured, in order to keep both the radiative transfer
model and the database tractable.

Finally, in sound-source localization, the acoustic input depends on both the source position, which can be observed \citep{TalmonCohenGannot2011,DeleforgeHoraud-MLSP12}, and of reverberations, that are strongly dependent on
the experimental conditions, and for which
ground-truth data are barely available.


\section{Gaussian Locally-Linear Mapping (GLLiM)}
\label{sec:gllim}


\color{black}
In this section, we describe in detail the GLLiM model which solves for inverse regression, \textit{i.e.}, the roles of input and response variables are exchanged such that the low-dimensional variable $\Xvect$ becomes the regressor.
GLLiM relies on a piecewise linear model in the following way.
Let $\{\xvect_n\}_{n=1}^{n=N} \in  \mathbb{R}^L$ and let us assume that
any realization
$(\yvect,\xvect)$ of $(\Yvect,\Xvect) \in \mathbb{R}^D \times \mathbb{R}^L$ is such that $\yvect$ is
the image of $\xvect$ by an affine transformation
$\tau_k$, among $K$, plus an error term. 
 This is modeled by a missing variable $Z$ such that
$Z=k$ if and only if $\Yvect$ is the image of
$\Xvect$ by $\tau_k$. The following decomposition of the joint probability distribution will be used:
\begin{align}
&p(\Yvect=\yvect,\Xvect=\xvect;\thetavect) =   \nonumber \\
\label{eq:joint-prob-distr}
& \sum_{k=1}^K p(\Yvect=\yvect | \Xvect=\xvect,Z=k;\thetavect) p(\Xvect=\xvect | Z=k;\thetavect) p(Z=k;\thetavect).
\end{align}
where $\thetavect$ denotes the vector of model parameters. The locally affine function that maps $\Xvect$ onto $\Yvect$ is:
\color{black}
\begin{equation}
\label{eq:model_yxz}
  \Yvect=\sum_{k=1}^K \mathbb{I}(Z=k) (\Amat_k \Xvect + \bvect_k +
  \Evect_{k})
\end{equation}
where $\mathbb{I}$ is the indicator function, matrix
$\Amat_k\in\mathbb{R}^{D\times L}$ and vector
$\bvect_k\in\mathbb{R}^D$ define the transformation $\tau_k$ and
$\Evect_{k}\in\mathbb{R}^D$ is an error term capturing both the
observation noise in $\mathbb{R}^D$ and the reconstruction error
due to the local affine approximation. Under the assumption that
$\Evect_{k}$ is a zero-mean Gaussian variable with covariance
matrix $\Sigmamat_k\in\mathbb{R}^{D\times D}$ that does not
depend on $\Xvect$, $\Yvect$,  and $Z$, we obtain:
\begin{equation}
 \label{eq:model_pYxz}
  p(\Yvect=\yvect|\Xvect=\xvect,Z=k;\thetavect) =
 \mathcal{N}(\yvect ;\Amat_k \xvect + \bvect_k ,\Sigmamat_k).
\end{equation}
To complete
the hierarchical definition of (\ref{eq:joint-prob-distr}) 
and enforce the affine
transformations to be local,  $\Xvect$ is assumed to follow a mixture of $K$ Gaussians defined by
\begin{align}
 p(\Xvect=\xvect |Z=k; \thetavect) & = \mathcal{N}(\xvect ; \cvect_k,\Gammamat_k)  \nonumber \\
 p(Z=k; \thetavect) &= \pi_k \label{eq:model_pXZ}
\end{align}
where $\cvect_k\in\mathbb{R}^L$, $\Gammamat_k\in\mathbb{R}^{L\times
L}$ and $\sum_{k=1}^K \pi_k=1$. 
\textcolor{black}{This model induces a partition of $\mathbb{R}^{L}$ into $K$ regions ${\cal R}_k$, where ${\cal R}_k$ is the region where the transformation $\tau_k$ is the most probable.}

It follows that the model parameters are: 
\begin{equation}
\label{eq:theta-def}
\thetavect = \{\cvect_k,\Gammamat_k,\pi_k,\Amat_k, \bvect_k,\Sigmamat_k\}_{k=1}^K.
\end{equation}
Once the parameter vector 
$\thetavect$ has been estimated, one obtains an inverse regression,  from $\mathbb{R}^L$ (low-dimensional space) to $\mathbb{R}^D$ (high-dimensional space), using 
the following \textit{inverse conditional density}:
\begin{equation}
 \label{eq:JGMM_forward_map}
 p(\Yvect=\yvect|\Xvect=\xvect;\thetavect) =
  \sum_{k=1}^K \frac{\pi_k\mathcal{N}(\xvect ; \cvect_k,\Gammamat_k)}{\sum_{j=1}^K\pi_j\mathcal{N}(\xvect ; \cvect_j,\Gammamat_j)} \mathcal{N}(\yvect;\Amat_k\xvect+\bvect_k,\Sigmamat_k).
\end{equation}
The forward regression, from $\mathbb{R}^D$ to $\mathbb{R}^L$  
is obtained from the \textit{forward conditional density}:
\begin{equation}
 \label{eq:JGMM_inverse_map}
 p(\Xvect=\xvect|\Yvect=\yvect;\thetavect) = \sum_{k=1}^K\frac{\pi_k^*\mathcal{N}(\yvect ; \cvect_k^*,\Gammamat_k^*)}{\sum_{j=1}^K\pi_j^*\mathcal{N}(\yvect ; \cvect_j^*.\Gammamat^*_j)}
 \mathcal{N}(\xvect;\Amat^*_k\yvect+\bvect^*_k,\Sigmamat_k^*).
\end{equation}
Notice that the above density is fully defined by $\thetavect$. Indeed, the \textit{forward parameter vector}: 
\begin{equation}
\label{eq:thetastar-def}
\thetavect^{*} = \{\cvect_k^{*},\Gammamat_k^{*},\pi_k^{*},\Amat_k^{*}, \bvect_k^{*},\Sigmamat_k^{*}\}_{k=1}^K
\end{equation}
 is obtained analytically using the following formulae:
\begin{align}
 \label{eq:GLLiMtoGLLimStar}
  \cvect_k^* & =\Amat_k\cvect_k+\bvect_k, \\
 \Gammamat_k^* & =\Sigmamat_k+\Amat_k\Gammamat_k\Amat_k\tp, \\
  \pi_k^* & = \pi_k,\\
  \Amat^*_k & = \Sigmamat_k^*\Amat_k\tp\Sigmamat_k^{-1},  \\
  \bvect^*_k & = \Sigmamat_k^*(\Gammamat_k^{-1}\cvect_k-\Amat_k\tp\Sigmamat_k^{-1}\bvect_k),\\
   \Sigmamat_k^* & = (\Gammamat_k^{-1}+\Amat_k\tp\Sigmamat_k^{-1}\Amat_k)^{-1}.
\end{align}
One interesting feature of the GLLiM model is that both densities (\ref{eq:JGMM_forward_map}) and (\ref{eq:JGMM_inverse_map}) are Gaussian mixtures parameterized by $\thetavect$. Therefore, one can use the expectation of (\ref{eq:JGMM_forward_map}) to obtain a low-to-high \textit{inverse regression function}:
\begin{equation}
 \label{eq:JGMM_forward_exp}
 \mathbb{E}[\Yvect=\yvect|\Xvect=\xvect;\thetavect]=\sum_{k=1}^K \frac{\pi_k\mathcal{N}(\xvect ; \cvect_k,\Gammamat_k)}{\sum_{j=1}^K\pi_j\mathcal{N}(\xvect ; \cvect_j,\Gammamat_j)}(\Amat_k\xvect+\bvect_k),
\end{equation}
or, even more interestingly, the expectation of (\ref{eq:JGMM_inverse_map}) to obtain a high-to-low \textit{forward regression function}:
\begin{equation}
 \label{eq:JGMM_inverse_exp}
 \mathbb{E}[\Xvect=\xvect|\Yvect=\yvect;\thetavect]=\sum_{k=1}^K \frac{\pi_k\mathcal{N}(\yvect ; \cvect_k^*,\Gammamat_k^*)}{\sum_{j=1}^K\pi_j\mathcal{N}(\yvect ; \cvect_j^*,\Gammamat^*_j)}(\Amat^*_k\yvect+\bvect^*_k).
\end{equation}

\subsection{Computational Tractability}

Let us analyze the cost of computing a low-to-high (inverse) regression. 
This computation relies on the estimation of the parameter vector $\thetavect$. Under the constraint that the $K$ transformations $\tau_k$ are affine, it is natural to assume that the error vectors $\Evect_k$ are modeled with  equal isotropic Gaussian noise, and hence we have $\{ \Sigmamat_k \}_{k=1}^{k=K} = \sigma^2\Imat_D$. The number of parameters to be estimated, \textit{i.e.}, the size of $\thetavect$, is  $K(1+L+DL+L(L+1)/2+D)$, for example it is equal to $30,060$ for $K=10$, $L=2$, and $D=1000$. 

If, instead, a high-to-low regression is directly estimated, the size of the parameter vector becomes $K(1+D+LD+D(D+1)/2+L)$, which is equal to $5,035,030$ in our example. In practice this is computationally intractable because it requires huge amounts of training data. Nevertheless, one may argue that the number of parameters could be drastically reduced by choosing covariance matrices $\{\Gammamat_k\}_{k=1}^{k=K}$ to be isotropic. However, this implies that an isotropic Gaussian mixture model is fitted to the high-dimensional data, which either would very poorly model the complexity of the data, or would require a large number of Gaussian components, leading to data over-fitting.

In Appendix \ref{app:JGMM} we show that
if $\thetavect$ is totally unconstrained, the joint distribution (\ref{eq:joint-prob-distr}) is that of an
unconstrained Gaussian mixture model (GMM) on the joint variable $[\Xvect; \Yvect]$, also referred to as \textit{joint GMM} (JGMM).
The symmetric roles of $\Xvect$ and $\Yvect$ in JGMM implies that low-to-high parameter estimation is strictly equivalent to high-to-low parameter estimation. However, JGMM requires the inversion of $K$ non-diagonal covariance matrices of size $(D+L)\times(D+L)$, which, again, becomes intractable for high-dimensional data.


\section{The Hybrid GLLiM Model}
\label{sec:plvm}
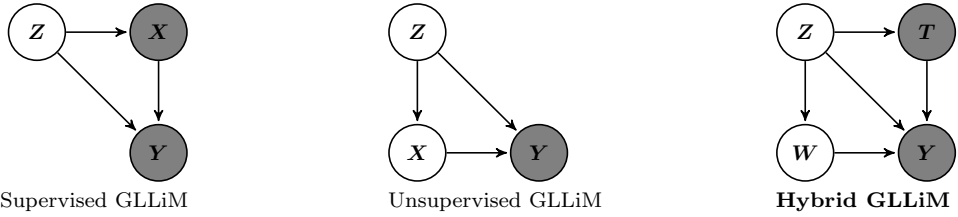
\begin{figure}[t!]
\centering
\begin{tabular}{p{0.3\linewidth}p{0.3\linewidth}p{0.35\linewidth}}\
\begin{tikzpicture}[->,>=stealth',shorten >=1pt,auto,node distance=1.6cm,
                    semithick]
  \tikzstyle{every state}=[circle]
  \node[state] (Z) {$\Zvect$};
  \node[state,fill=gray] (T) [right of=Z] {$\Xvect$};
  \node[state,fill=gray] (Y) [below of=T] {$\Yvect$};
  \path (Z) edge (T)
    (T) edge (Y)
    (Z) edge (Y);
\end{tikzpicture} &
\begin{tikzpicture}[->,>=stealth',shorten >=1pt,auto,node distance=1.6cm,
                    semithick]
  \tikzstyle{every state}=[circle]
  \node[state] (Z) {$\Zvect$};
  \node[state] (W) [below of=Z]{$\Xvect$};
  \node[state,fill=gray] (Y)  [right of=W] {$\Yvect$};
  \path (Z) edge (W)
    (W) edge (Y)
    (Z) edge (Y);
\end{tikzpicture} &
\begin{tikzpicture}[->,>=stealth',shorten >=1pt,auto,node distance=1.6cm,
                    semithick]
  \tikzstyle{every state}=[circle]
  \node[state] (Z) {$\Zvect$};
  \node[state] (W) [below of=Z] {$\Wvect$};
  \node[state,fill=gray] (T) [right of=Z] {$\Tvect$};
  \node[state,fill=gray] (Y) [right of=W] {$\Yvect$};
  \path (Z) edge (W)
        (Z) edge (T)
    (T) edge (Y)
    (W) edge (Y)
    (Z) edge (Y);
\end{tikzpicture} \\
 Supervised GLLiM 
 & Unsupervised GLLiM 
 & \textbf{Hybrid GLLiM} 
\end{tabular}
\caption{Graphical representation of the GLLiM models. White-filled circles correspond to unobserved variables while grey-filled circles correspond to observed variables.}
\label{fig:graphical_model}
\end{figure}
The model just described can be learned with standard EM inference methods if $\Xvect$ and $\Yvect$ are both observed.
The key idea in this paper is to treat $\Xvect$ as a \textit{partially-latent} variable, namely 
$$
\Xvect=\left[
  \begin{array}{c}
   \Tvect \\
   \Wvect
  \end{array}
 \right],
$$ where $\Tvect\in\mathbb{R}^{L_{\textrm{t}}}$ is observed and
$\Wvect\in\mathbb{R}^{L_{\textrm{w}}}$ is latent
($L=L_{\textrm{t}}+L_{\textrm{w}}$). Graphical representations of supervised GLLiM, unsupervised GLLiM, and hybrid GLLiM models are illustrated in Figure~\ref{fig:graphical_model}.
In hybrid GLLiM, the estimation of model parameters uses observed pairs 
$\{\yvect_n,\tvect_n\}_{n=1}^N$ while it must also be constrained by the presence of the latent variable $\Wvect$. 
This can be seen as a \textit{latent-variable augmentation} of classical regression, where
the observed realizations of $\Yvect$ are affected by
the unobserved variable $\Wvect$. It can also be viewed as a
variant of dimensionality reduction since the unobserved
low-dimensional variable $\Wvect$ must be recovered from
$\{(\yvect_n,\tvect_n)\}_{n=1}^N$. The
decomposition of $\Xvect$  into observed and latent parts implies that some of the model
parameters must be decomposed as well, namely $\cvect_k$,
$\Gammamat_k$ and $\Amat_k$. Assuming the independence of $\Tvect$
and $\Wvect$ given $Z$ we write:
\begin{equation}
 \label{eq:SOLVA_decomp}
 \cvect_k=\left[
  \begin{array}{c}
   \cvect^{\textrm{t}}_{k}\\
   \cvect^{\textrm{w}}_{k}
  \end{array}
 \right],\;
 \Gammamat_k=\left[
  \begin{array}{lc}
    \Gammamat_k^{\textrm{t}}     & \zerovect \\
    \zerovect & \Gammamat_k^{\textrm{w}}
  \end{array}
 \right],\;
 \Amat_k= \left[ \begin{array}{cc} \Amat_k^{\textrm{t}} & \Amat_k^{\textrm{w}} \end{array} \right].
\end{equation}
It follows that (\ref{eq:model_yxz}) rewrites as
\begin{equation}
\label{eq:model_ytwz}
  \Yvect=\sum_{k=1}^K \mathbb{I}(Z=k) (\Amat_k^{\textrm{t}} \Tvect + \Amat_k^{\textrm{w}} \Wvect + \bvect_k +
  \Evect_{k})
\end{equation}
or equivalently:
\begin{equation}
\label{eq:model_ytwzbis}
  \Yvect=\sum_{k=1}^K \mathbb{I}(Z=k) (\Amat_k^{\textrm{t}} \Tvect + \bvect_k + \Amat_k^{\textrm{w}} \cvect_k^{\textrm{w}} +
  \Evect'_{k})
\end{equation}
where the error vector $\Evect'_k$ is modeled by a zero-centered Gaussian variable
with a ${D\times D}$ covariance matrix given by
\begin{equation}
\label{eq:sigma-prime}
\Sigmamat'_k=\Sigmamat_k+\Amat_k^{\textrm{w}}\Gammamat_k^{\textrm{w}}\Amat_k^{\textrm{w}\top}.
\end{equation}
Considering realizations of variables $\Tvect$ and $\Yvect$, one
may thus view hybrid GLLiM as a supervised GLLiM model in which the noise
covariance has an unconventional structure, namely (\ref{eq:sigma-prime}), where
$\Amat_k^{\textrm{w}}\Gammamat_k^{\textrm{w}}\Amat_k^{\textrm{w}\top}$
is at most a rank-$L_{\textrm{w}}$ matrix. When $\Sigmamat_k$ is diagonal,  this
structure is that of factor analysis with at most $L_{\textrm{w}}$ factors, and
represents a flexible compromise between a full covariance with
$O(D^2)$ parameters on one side, and a diagonal covariance with $O(D)$
parameters on the other side.  Let us consider the isotropic case, {\it i.e.},
$\Sigmamat_k=\sigma_k^2\Imat_D ,$ for all $k=1: K$. We obtain the following three cases for the proposed model:
\begin{itemize}
\item
$L_{\textrm{w}}=0$. This is the fully supervised case, $\Sigmamat'_k= \Sigmamat_k$, and is equivalent to the mixture of local linear experts (MLE) model \citep{XuJordanHinton95}.
\item $L_{\textrm{w}}=D$. $\Sigmamat'_k$ takes the form of a general covariance matrix
and we obtain the JGMM model \citep{KainMacon98,QiaoMinematsu09} (see Appendix \ref{app:JGMM} for a proof). This is the most general GLLiM model, which requires
 the estimation of $K$ full covariance matrices of size 
${(D+L)\times (D+L)}$. This model becomes over-parameterized and intractable in high dimensions.
\item $0< L_{\textrm{w}} < D$. This corresponds to the hybrid GLLiM model, and yields a wide variety of
novel regression models \textit{in between} MLE and JGMM.
\end{itemize}
In Section~\ref{sec:results}, we experimentally show that in some practical cases it is advantageous to use hybrid GLLiM, \textit{i.e.}, the response variable is only partially observed during training, yielding better results than with MLE, JGMM, or a number of state of the art regressions techniques.

As summarized in Table \ref{tab:unif}, a number of existing methods can be seen as particular instances of hybrid GLLiM where either $L_{\textrm{t}}$ or $L_{\textrm{w}}$ is equal to $0$. Several regression models ($L_\textrm{w}=0$) are instances of hybrid GLLiM, \textit{i.e.}, supervised GLLiM. This is the case for
the mixture of local linear experts (MLE) \citep{XuJordanHinton95} where the noise covariances $\{\Sigmamat_k\}_{k=1}^K$ are isotropic. Probabilistic piecewise affine mapping (PPAM) \citep{DeleforgeHoraud-MLSP12} may be viewed as a variant of MLE where $\{\Gammamat_k\}_{k=1}^K$ have equal determinants. As already mentioned, it is shown in Appendix~\ref{app:JGMM} that JGMM \citep{KainMacon98,QiaoMinematsu09} corresponds to the case of unconstrained parameters. The mixture of linear regressors (MLR) \citep{quandt1978estimating,deVeaux89,jedidi1996estimating} may also be viewed,  as a supervised GLLiM model where covariances $\{\Gammamat_k\}_{k=1}^K$ are set to $\eta\Imat_L$ with $\eta\rightarrow\infty$, i.e., there is no prior on $\Xvect$. Similarly, several dimensionality reduction models ($L_\textrm{t}=0$) are instances of hybrid GLLiM, \textit{i.e.}, unsupervised GLLiM. This is the case for probabilistic principal component analysis (PPCA) \citep{tipping1999probabilistic} and its mixture version (MPPCA) \citep{TippingBishop99} where the noise covariances $\{\Sigmamat_k\}_{k=1}^K$ are isotropic. Mixture of factor analyzers (MFA) \citep{GhahramaniHinton96} corresponds to diagonal noise covariances, probabilistic canonical correlation analysis (PCCA) \citep{bach2005probabilistic} corresponds to block-diagonal noise covariances, and residual component analysis (RCA) \citep{kalaitzis2011residual} corresponds to fixed (not estimated) noise covariances. The generative topographic mapping (GTM) \citep{Bishop98} may also be viewed as an unsupervised GLLiM model where covariances $\{\Gammamat_k\}_{k=1}^K$ are set to $\epsilon\Imat_L$ with $\epsilon\rightarrow0$, \textit{i.e.}, the prior on $\Xvect$ is a mixture of Dirac functions.
While providing a unifying perspective over these methods, hybrid GLLiM enables a wide range of generalizations corresponding to $L_{\textrm{t}}>0$, $L_{\textrm{w}}>0$. 

\begin {table}[t!]
   \center
   \begin{tabular}{l|llllll|lll}
   Method  & $\cvect_k$ & $\Gammamat_k$ & $\pi_k$ & $\Amat_k$ & $\bvect_k$ & $\Sigmamat_k$
           & $L_{\textrm{t}}$ & $L_{\textrm{w}}$ & $K$ \\
   \hline
   MLE \citep{XuJordanHinton95}      & - & - & - & - & - & diag & - & 0 & - \\
   MLR  \citep{jedidi1996estimating} & $\zerovect_L $ & $\infty\Imat_L$ & -     & -     & -              & iso+eq   & - & $0$ & -   \\
   JGMM \citep{QiaoMinematsu09} & -              & -               & -     & -     & -              & -          & - & $0$ & -   \\
   PPAM \citep{DeleforgeHoraud-MLSP12} & -              & $|$eq$|$       & eq   & -     & -              & diag+eq  & - & $0$ & -   \\
   \hline
   GTM  \citep{Bishop98} & fixed          & $\zerovect_L$   & eq.   & eq.   & $\zerovect_D $ & iso+eq & $0$ & - & -   \\
   PPCA \citep{tipping1999probabilistic} & $\zerovect_L $ & $\Imat_L$       & -     & -     & -              & iso     & $0$ & - & $1$ \\
   MPPCA \citep{TippingBishop99} & $\zerovect_L $ & $\Imat_L$       & -     & -     & -              & iso     & $0$ & - & -   \\
   MFA  \citep{GhahramaniHinton96} & $\zerovect_L $ & $\Imat_L$       & -     & -     & -              & diag    & $0$ & - & -   \\
   PCCA \citep{bach2005probabilistic} & $\zerovect_L $ & $\Imat_L$       & -     & -     & -              & block     & $0$ & - & $1$ \\
   RCA   \citep{kalaitzis2011residual}  & $\zerovect_L $ & $\Imat_L$       & -     & -     & -              & fixed     & $0$ & - & $1$ \\
   \end{tabular}
   \caption{\label{tab:unif} 
   \color{black}
   This table summarizes the link between the proposed model and several existing methods. The first three rows corresponds to supervised GLLiM methods ($L_{\textrm{w}}=0$, Fig.~\ref{fig:graphical_model}(a)) while the last six rows correspond to unsupervised GLLiM methods ($L_{\textrm{t}}=0$, Fig.~\ref{fig:graphical_model}(b)). The following symbols are used: ``diag" (diagonal covariance matrices), ``eq" (equal covariance matrices), ``$|$eq$|$" (equal determinants), ``fixed" (not estimated), ``block" (block-diagonal covariance matrices), ``-" (unconstrained).
   \color{black}
   }
\end{table}
\color{black}

Finally, it is worth to be noticed that an appropriate choice of the kernel function in the Gaussian process latent variable model (GPLVM) \citep{lawrence2005} allows to account for a partially
observed input variable. This was notably studied in \citep{fusi2012joint}. However, as explained in \citep{lawrence2005}, the mapping yielded by GPLVM cannot be ``inverted", due to the non-linear nature of the kernels used in practice. Hence, GPLVM allows regression with partially-latent \textit{input}, and not with partially-latent \textit{response}. The existence of a closed-form expression for the forward regression function, \textit{i.e.}, (\ref{eq:JGMM_inverse_exp}), is therefore a crucial ingredient of the proposed model that fully justifies the usefulness of GLLiM when the task is to regress high-dimensional data onto a partially-observed response.

\section{Expectation-Maximization for Hybrid-GLLiM}
\label{sec:algo}  In this section we devise an EM algorithm to
estimate the parameters of the proposed model. The principle of the suggested
algorithm is based on a data augmentation strategy that consists
of augmenting the observed variables with the unobserved ones, in
order to facilitate the subsequent maximum-likelihood search over
the parameters.
\subsection{Data Augmentation Schemes}
 There are two sets of missing variables, $Z_{1:N}=
\{Z_n\}_{n=1}^N$ and $\Wvect_{1:N}=\{\Wvect_n\}_{n=1}^N$,
associated with the training data set $(\yvect,
\tvect)_{1:N}=\{\yvect_n,\tvect_n\}_{n=1}^N$, given the number $K$
of linear components and the latent dimension $L_\textrm{w}$. Two
augmentation schemes arise naturally. The first scheme is referred
to as general hybrid GLLiM-EM, or general-hGLLiM, and consists of augmenting the observed data
with both variables $(Z,\Wvect)_{1:N}$ while the second scheme,
referred to as marginal-hGLLiM, consists of integrating out the
continuous variables $\Wvect_{1:N}$ previous to data augmentation
with the discrete variables $Z_{1:N}$. \textcolor{black}{The
difference between these two schemes is in the amount of missing
information and this may be of interest considering the well-known
fact that the convergence rates of EM procedures are determined by
the portion of missing information in the complete data. To accelerate
standard EM algorithms it is natural to decrease the amount of
missing data, but the practical computational gain is effective
only on the premise that the corresponding M-step can be solved
efficiently. Another strategy, as a suitable tradeoff between
simplicity (or efficiency) and convergence, is based on an
extension of the Expectation Conditional Maximization (ECM)
algorithm \citep{MengRubin1993}, referred to as the Alternating
ECM (AECM) algorithm \citep{MengDyk1997}. In AECM, the amount of
missing data is allowed to be different in each conditional
maximization  (CM) step. An application of AECM to mixture of factor analysers (MFA) with all
its CM-steps in closed-form is given in \citep{McLachlan2003} and
can be compared to the standard EM for MFA described in
\citep{GhahramaniHinton96}. In the case of the proposed hybrid GLLiM
model, as it is the case for MFA, using an AECM algorithm
typically affects the estimations of the Gaussian means, namely
the $\bvect_k$'s in (\ref{eq:model_pYxz}). For the latter
estimations, the expected empirical weighted mean of the
observations is not recovered with standard EM while it is with
AECM (see details in Section \ref{subsec:algo}).}

\subsection{Generalization of Other Algorithms}
The general hybrid GLLiM-EM algorithm, described in detail below, leads to
closed-form expressions for a wide range of constraints onto the
covariance matrices $\{\Gammamat_k\}_{k=1}^K$ and
$\{\Sigmamat_k\}_{k=1}^K$. Moreover, the algorithm can be applied
to both supervised ($L_\textrm{w}=0$) and unsupervised
($L_\textrm{t}=0$) GLLiM models. Hence, it can be viewed as a
generalization of a number of EM inference techniques for
regression, {\it e.g.}, MLR, MLE, JGMM, GTM, or for dimensionality
reduction, {\it e.g.}, MPPCA, MFA, PPCA, and RCA. The marginal
hybrid GLLiM-EM algorithm, which is described in detail in
Appendix~\ref{app:marginal}, is less general. Nevertheless, it is
of interest because it provides both an algorithmic insight into
the hybrid GLLiM model as well as a natural initialization strategy for
the general algorithm. Note that, as
mentioned  in Appendix~\ref{app:marginal}, the marginal hybrid GLLiM-EM
also admits an ECM variant.
\textcolor{black}{A comprehensive Matlab toolbox including all necessary functions for GLLiM as well as illustrative examples is available online\footnote{\black{https://team.inria.fr/perception/gllim\_toolbox/}}.}
\subsection{Non-Identifiability Issues}
Notice that the means $\{\cvect_k^{\textrm{w}}\}_{k=1}^K$ and
covariance matrices $\{\Gammamat_k^{\textrm{w}}\}_{k=1}^K$ must be
fixed to avoid non-identifiability issues. Indeed, changing their
values respectively corresponds to shifting and scaling the
unobserved variables $\Wvect_{1:N}\in
\mathbb{R}^{L_{\textrm{w}}}$, which can be compensated by
changes in the parameters of the affine transformations
$\{\Amat_k^{\textrm{w}}\}_{k=1}^K$ and $\{\bvect_k\}_{k=1}^K$. The
same issue is observed in all latent variable models used for
dimensionality reduction and is always solved by fixing these
parameters. In GTM \citep{Bishop98} the means are spread on a
regular grid and the covariance matrices are set to $\zerovect$
(Dirac functions), while in MPPCA \citep{TippingBishop99} and MFA
\citep{GhahramaniHinton96} all means and covariance matrices are
respectively set to zero and to identity matrices. The latter option will also be used in our experiments (sections~\ref{subsec:synth}, \ref{subsec:pose} and \ref{subsec:planet}), but
for the sake of generality, the
following EM algorithm is derived for means and
covariance matrices that are arbitrarily fixed.

\subsection{The General Hybrid GLLiM-EM Algorithm}
\label{subsec:algo} Considering the complete data, with $(\Yvect,
\Tvect)_{1:N}$ being the observed variables and  $(Z, \Wvect)_{1:N}$
being the missing ones, the corresponding EM algorithm consists of
estimating the parameter vector $\thetavect^{(i+1)}$ that maximizes the expected complete-data log-likelihood, given the current parameter vector $\thetavect^{(i)}$ and the observed data:
\begin{equation}
\label{eq:expected-completedata}
\thetavect^{(i+1)} = \arg\max\limits_{\thetavect}
\mathbb{E}[\log p((\yvect,\tvect,\Wvect,Z)_{1:N};\thetavect) | (\yvect,\tvect)_{1:N} ;\thetavect^{(i)}].
\end{equation}
Using that $\Wvect_{1:N}$ and $\Tvect_{1:N}$ are independent
conditionally on $Z_{1:N}$ and that
$\{\cvect_k^{\textrm{w}}\}_{k=1}^K$ and
$\{\Gammamat_k^{\textrm{w}}\}_{k=1}^K$ are fixed, maximizing (\ref{eq:expected-completedata})
is then equivalent to maximizing the following expression:
\begin{equation} \mathbb{E}_{r_{Z}^{(i+1)}}[\mathbb{E}_{r_{W|Z}^{(i+1)}}[\log
p(\yvect_{1:N}\; | \; (\tvect,\Wmat,Z)_{1:N};\thetavect)]
 + \log
p((\tvect,Z)_{1:N};\thetavect)] \label{eq:Q-PSM-EM}
\end{equation}
where $r_{Z}^{(i+1)}$ and $r_{W|Z}^{(i+1)}$ denote
the posterior distributions
\begin{align}
\label{eq:r_z}
r_{Z}^{(i+1)} &= p(\Zmat_{1:N} | (\yvect,\tvect)_{1:N} ;\thetavect^{(i)}) \\
\label{eq:r_wz}
r_{W|Z}^{(i+1)} &= p(\Wmat_{1:N} |(\yvect,\tvect,Z)_{1:N};\thetavect^{(i)}).
\end{align}
It follows that the E-step splits
into an \textbf{E-W} step and an \textbf{E-Z} step in the following way. For the
sake of readability, the current iteration superscript $(i+1)$ is
replaced with a tilde. Hence, $\thetavect^{(i+1)} = \widetilde{\thetavect}$ (the model parameter vector).

\begin{description}
\item[\textbf{E-W-step:}]
\label{par:SOLVA_EWstep} The posterior probability
$\widetilde{r}_{W|Z}$, given parameter estimates, is fully
defined by computing the distributions
$p(\wvect_n|Z_n=k,\tvect_n,\yvect_n;\thetavect^{(i)})$, for all $n$ and all $k$, which can be
shown to be Gaussian, with mean $\widetilde{\muvect}_{nk}^{\textrm{w}}$ and covariance matrix
$\widetilde{\Smat}_k^{\textrm{w}}$ given by:
\begin{align}
 \label{eq:SOLVA_EWstep3}
 \widetilde{\muvect}_{nk}^{\textrm{w}}&= \widetilde{\Smat}_k^{\textrm{w}}\big((\Amat_k^{\textrm{w}(i)})^\top(\Sigmamat_k^{(i)})\inverse(\yvect_n-\Amat_k^{\textrm{t}(i)}\tvect_n-\bvect_k^{(i)}) + (\Gammamat_k^{\textrm{w}(i)})\inverse \cvect_k^{\textrm{w}(i)}\big)\\
 \widetilde{\Smat}_k^{\textrm{w}}&= \big((\Gammamat_k^{\textrm{w}(i)})\inverse +(\Amat_k^{\textrm{w}(i)})^\top(\Sigmamat_k^{(i)})\inverse \Amat_k^{\textrm{w}(i)}\big)\inverse
\end{align}

\textcolor{black}
Conditionally to $Z_n=k$, equation
(\ref{eq:model_ytwzbis}) shows that this step amounts to a factor
analysis step. Indeed, we recover standard formula for the
posterior over latent factors where the observations are replaced
by the {\it current residuals}, namely
$\yvect_n-\Amat_k^{\textrm{t}(i)}\tvect_n-\bvect_k^{(i)}$. 
\textcolor{black}{Details on the \textbf{E-W-step} are given in Section 1.1 of the Supplementary Materials.}

\item[\textbf{E-Z-step:}] \label{par:SOLVA_EZstep} The posterior
probability $\widetilde{r}_{Z}$ is defined by:
\begin{equation}
 \label{eq:SOLVA_EZstep}
 \widetilde{r}_{nk}=p(Z_n=k|\tvect_n,\yvect_n;\thetavect^{(i)}) = \frac{\pi_k^{(i)}p(\yvect_n,\tvect_n|Z_n=k;\thetavect^{(i)})}{\textstyle\sum_{j=1}^K\pi_j^{(i)}p(\yvect_n,\tvect_n|Z_n=j;\thetavect^{(i)})}
\end{equation}
for all $n$ and all $k$, where
$$
p(\yvect_n,\tvect_n|Z_n=k;\thetavect^{(i)})= p(\yvect_n|\tvect_n,
Z_n=k;\thetavect^{(i)})\; p(\tvect_n|Z_n=k;\thetavect^{(i)}).
$$
The
second term is equal to
$\mathcal{N}(\tvect_n;\cvect_k^{\textrm{t}},\Gammamat_k^{\textrm{t}})$
by virtue of (\ref{eq:model_pXZ}) and (\ref{eq:SOLVA_decomp}) while it is
clear from (\ref{eq:model_ytwzbis})  that
$$
p(\yvect_n|\tvect_n,
Z_n=k;\thetavect^{(i)}) =
\mathcal{N}(\yvect_n;\Amat_k^{(i)}[\tvect_n ;
\cvect_k^{\textrm{w}}]+\bvect_k^{(i)},
\Amat_k^{\textrm{w}(i)}\Gammamat_k^{\textrm{w}}\Amat_k^{\textrm{w}(i)\top}
+ \Sigmamat_k^{(i)} ).
$$
The maximization of (\ref{eq:expected-completedata}) can then be
performed using the posterior probabilities $\widetilde{r}_{nk}$ and
the sufficient statistics $\widetilde{\muvect}_{nk}^{\textrm{w}}$
and $\widetilde{\Smat}_k^{\textrm{w}}$. We use the following
notations: $\widetilde{r}_k=\sum_{n=1}^N\widetilde{r}_{nk}$ and
$\widetilde{\xvect}_{nk}=[\tvect_n;\widetilde{\muvect}_{nk}^{\textrm{w}}]\in\mathbb{R}^L$.
It can be easily seen from the decomposition (\ref{eq:Q-PSM-EM}) of
(\ref{eq:expected-completedata}), that the M-step can be divided into two separate steps.

First, the updates of parameters $\widetilde{\pi}_k,
\widetilde{\cvect}_k^{\textrm{t}}$ and
$\widetilde{\Gammamat}_k^{\textrm{t}}$ correspond to those of a standard Gaussian mixture model
on $\Tvect_{1:N}$, so that we get straightforwardly:

\item[\textbf{M-GMM-step:}]
\begin{align}
\widetilde{\pi}_k &=\frac{\widetilde{r}_k}{N} \\
\widetilde{\cvect}_k^t &= \sum_{n=1}^N\frac{\widetilde{r}_{kn}}{\widetilde{r}_{k}}\tvect_n,\\
\widetilde{\Gammamat}_k^t &= \sum_{n=1}^N\frac{\widetilde{r}_{kn}}{\widetilde{r}_{k}}(\tvect_n-\widetilde{\cvect}_k^t)(\tvect_n-\widetilde{\cvect}_k^t)\tp.
\end{align}
Second, the updating of the mapping parameters $\{\Amat_k, \bvect_k, \Sigmamat_k\}_{k=1}^K$ is also in closed-form.
\textcolor{black}{Details on the \textbf{M-mapping-step} below are provided in Section 1.2 of the Supplementary Materials.} 

\item[\textbf{M-mapping-step:}]
\begin{equation}
\label{eq:Ak_up}
\widetilde{\Amat}_k=\widetilde{\Ymat}_k\widetilde{\Xmat}_k\tp(\widetilde{\Smat}_k^{\textrm{x}}+\widetilde{\Xmat}_k\widetilde{\Xmat}_k\tp)^{-1}
\end{equation}
where:
\begin{align}
\widetilde{\Smat}_k^{\textrm{x}} &=\left[
  \begin{array}{ll}
    \zerovect & \zerovect\\
    \zerovect & \widetilde{\Smat}_k^{\textrm{w}}
  \end{array}
 \right], \\
 \widetilde{\Xmat}_k &
= \frac{1}{\sqrt{\widetilde{r}_k}}
 \left[
\begin{array}{ccc}
\sqrt{\widetilde{r}_{1k}}(\widetilde{\xvect}_{1k}-\widetilde{\xvect}_k) & \dots  & \sqrt{\widetilde{r}_{Nk}}
 (\widetilde{\xvect}_{Nk}-\widetilde{\xvect}_k)
 \end{array}
\right], \\
%
 \widetilde{\Ymat}_k &
 = \frac{1}{\sqrt{\widetilde{r}_k}}
 \left[
\begin{array}{ccc}
\sqrt{\widetilde{r}_{1k}}(\yvect_1-\widetilde{\yvect}_k) & \dots & \sqrt{\widetilde{r}_{Nk}}(\yvect_N-\widetilde{\yvect}_k)
\end{array}
\right], \\
 \color{black}
 \widetilde{\xvect}_k &= \sum_{n=1}^N \frac{\widetilde{r}_{kn}}{\widetilde{r}_k} \widetilde{\xvect}_{nk},\\
 \widetilde{\yvect}_k &= \sum_{n=1}^N \frac{\widetilde{r}_{kn}}{\widetilde{r}_k} \yvect_n.
\end{align}
When $L_\textrm{w}=0$ then $\widetilde{\Smat}_k^{\textrm{x}}=0$
and the expression (\ref{eq:Ak_up}) of $\widetilde{\Amat}_k$ is that of
standard linear regression from $\{\tvect_n\}_{n=1}^N$ to
$\{\yvect_n\}_{n=1}^N$ weighted by
$\{\widetilde{r}_{nk}\}_{n=1}^N$. When $L_\textrm{t}=0$ then
$\widetilde{\Smat}_k^{\textrm{x}}=\widetilde{\Smat}_k^{\textrm{w}}$
and we obtain the principal components update of the EM
algorithm for PPCA \citep{tipping1999probabilistic}. The intercept
parameter is updated with:
\begin{equation}
 \widetilde{\bvect}_k =
 \sum_{n=1}^N\frac{\widetilde{r}_{kn}}{\widetilde{r}_{k}}(\yvect_n-\widetilde{\Amat}_k\widetilde{\xvect}_{nk}),
\end{equation}
or equivalently:
\[
 \widetilde{\bvect}_k =
 \sum_{n=1}^N \frac{\widetilde{r}_{kn}}{\widetilde{r}_{k}} (\yvect_n-\widetilde{\Amat}_k^{\textrm{t}}
 \tvect_n) - \widetilde{\Amat}_k^{\textrm{w}}
 \sum_{n=1}^N \frac{\widetilde{r}_{kn}}{\widetilde{r}_{k}} \widetilde{\muvect}_{nk}^{\textrm{w}}.
 \]
The second term in this expression is the one that would
disappear in an AECM algorithm.
Finally, we obtain the following expression
for $\widetilde{\Sigmamat}_k$:
\begin{equation}
 \label{eq:SOLVA_Sigmak}
 \widetilde{\Sigmamat}_k = \widetilde{\Amat}_k^{\textrm{w}}\widetilde{\Smat}_k^{\textrm{w}}\widetilde{\Amat}_k^{\textrm{w}\top}
 +
 \sum_{n=1}^N\frac{\widetilde{r}_{kn}}{\widetilde{r}_{k}}(\yvect_n-\widetilde{\Amat}_k\widetilde{\xvect}_{nk}- \widetilde{\bvect}_k)
 (\yvect_n-\widetilde{\Amat}_k\widetilde{\xvect}_{nk}- \widetilde{\bvect}_k)\tp.
\end{equation}
\end{description}
Note that the previous formulas can be seen as
standard ones after {\it imputation} of the missing variables
$\wvect_n$ by their mean values
$\widetilde{\muvect}_{nk}^{\textrm{w}}$ via the definition of
$\tilde{\xvect}_{nk}$. As such a direct imputation by the mean
necessarily underestimates the variance, the above formula also
contains an additional term typically involving the variance
$\tilde{\Svect}_k^{\textrm{w}}$ of the missing data.

 Formulas are
given for unconstrained parameters, but can be straightforwardly
adapted to different constraints. For instance, if
$\{\Mmat_k\}_{k=1}^K\subset\mathbb{R}^{P\times P}$ are solutions
for unconstrained covariance matrices $\{\Sigmamat_k\}_{k=1}^K$ or
$\{\Gammamat_k\}_{k=1}^K$, then solutions with diagonal (diag),
isotropic (iso) and/or equal  (eq) for all $k$ constraints are
respectively given by
$\Mmat_k^{\textrm{diag}}=\operatorname{diag}(\Mmat_k)$,
$\Mmat_k^{\textrm{iso}}=\operatorname{tr}(\Mmat_k)\Imat_P\black{/P}$
and $\Mmat^{\textrm{eq}}=\sum_{k=1}^K\widetilde{\pi}_k\Mmat_k$.

\subsection{Algorithm Initialization}
In general, EM algorithms are known to be
sensitive to initialization and likely to converge to
local maxima of the likelihood, if not appropriately initialized.
Initialization could be achieved either by choosing a set of parameter
values and proceeding with the E-step, or by choosing a set of posterior
probabilities and proceeding with the M-step. The
general
hybrid GLLiM-EM algorithm
however, is such that there is no straightforward way of choosing
a complete set of initial posteriors (namely $r_{nk}^{(0)},
\muvect_{nk}^{\textrm{w}(0)}$ and $\Smat_k^{\textrm{w}(0)}$ for
all $n,k$) or a complete set of initial parameters
$\thetavect^{(0)}$ including all the affine transformations.
This issue is addressed by deriving the \textit{marginal} hybrid GLLiM-EM algorithm, a variant of the%
general hybrid GLLiM-EM, in which latent variables
$\Wvect_{1:N}$ are integrated out, leaving only the estimation of
posteriors $r_Z$ in the E-step. Full details on this variant are
given in Appendix \ref{app:marginal}. As explained there, this
variant is much easier to initialize but  it has closed-form steps
only if the covariance matrices $\{\Sigmamat_k\}_{k=1}^K$ are
isotropic and distinct. In practice, we start with one iteration of
the marginal hybrid GLLiM-EM to obtain a set of initial parameters
$\theta^{(0)}$ and continue with the general hybrid GLLiM-EM until convergence.

\subsection{\black{Latent Dimension Estimation Using BIC}}
\black{
Once a set of parameters $\widetilde{\thetavect}$ has been learned with hGLLiM-EM, the \textit{Bayesian information criterion} (BIC) can be computed as follows:
\begin{equation}
 BIC(\widetilde{\thetavect},N) = -2 \mathcal{L}(\widetilde{\thetavect}) + \mathcal{D}(\widetilde{\thetavect})\log N
\end{equation}
where $\mathcal{L}$ denotes the observed-data log-likelihood and $\mathcal{D}(\widetilde{\thetavect})$ denotes the dimension of the complete parameter vector $\widetilde{\thetavect}$. Assuming, \textit{e.g.}, isotropic and equal noise covariance matrices $\{\Sigmamat_k\}_{k=1}^K$, we have:
\begin{align}
 \mathcal{D}(\widetilde{\thetavect}) &= K(D(L_{\textrm{w}}+L_{\textrm{t}}+1)+L_{\textrm{t}}(L_{\textrm{t}}+3)/2+1)  \hspace{2mm} \textrm{and} \\
 \mathcal{L}(\widetilde{\thetavect}) &= \sum_{n=1}^N\log p(\yvect_n,\tvect_n;\widetilde{\thetavect})
\end{align}
where $p(\yvect_n,\tvect_n;\widetilde{\thetavect})$ is given by the denominator of (\ref{eq:SOLVA_EZstep}). A natural way to choose a value for $L_{\textrm{w}}$ is, for a given value of $K$, to train hGLLiM-EM with different values of $L_{\textrm{w}}$, and select the value minimizing BIC. We will refer to the corresponding method as hGLLiM-BIC. It has the advantage of not requiring the parameter $L_\textrm{w}$, but it is more computationally demanding because it requires to run hGLLiM-EM for all tested values of $L_{\textrm{w}}$. However, efficient implementations could parallelize these runs.}

\section{Experiments and Results}
\label{sec:results}

\subsection{\black{Evaluation methodology}}
\label{subsec:xp_method}
\black{In this Section, we evaluate the performance of the general hybrid GLLiM-EM algorithm (hGLLiM) proposed in Section \ref{subsec:algo} on 3 different datasets. In Section \ref{subsec:synth} we inverse high-dimensional functions using synthetic data. In Section \ref{subsec:pose} we retrieve pose or light information from face images. In Section \ref{subsec:planet} we recover some physical properties of the Mars surface from hyperspectral images. For each of these 3 datasets we consider situations where the target low-dimensional variable is only partially observed during training. hGLLiM-BIC and other hGLLiM models corresponding to a fixed value of $L_{\textrm{w}}$ are tested. The latter are denoted hGLLiM-$L_{\textrm{w}}$. As mentioned in Table \ref{tab:unif}, hGLLiM-0 is actually equivalent to a mixture of local linear experts model \citep{XuJordanHinton95} and will thus be referred to as MLE in this Section. In practice, the MLE parameters are estimated using the proposed general hGLLiM-EM algorithm, by setting $L_{\textrm{w}}$ to $0$. In all tasks considered, $N$ observed training couples $\{(\tvect_n,\yvect_n)\}_{n=1}^{N}$ are used to obtain a set of parameters. Then, we use the forward mapping function (\ref{eq:JGMM_inverse_exp}) to compute an estimate $\hat{\tvect}'$ given a test observation $\yvect'$ (please refer to Section \ref{sec:gllim}). This is repeated for $N'$ test observations $\{\yvect'_n\}_{n=1}^{N'}$. The training and the test sets are disjoints in all experiments.
Note that MLE was not developed in the context of inverse-then-forward regression in its original paper, and hence, was not suited for high-to-low dimensional regression. Recently, \citep{DeleforgeHoraud-MLSP12} combined a variant of MLE with an inverse-then-forward strategy. This variant is called PPAM and includes additional constraints on $\{\Gammamat_k\}_{k=1}^{k=K}$ (see Table \ref{tab:unif}).}

Hybrid GLLiM and MLE models are also compared to three other regression techniques, namely joint GMM (JGMM) \citep{QiaoMinematsu09} which
is equivalent to hGLLiM with $L_\textrm{w}\ge D$ (see Section
\ref{sec:plvm} and Appendix \ref{app:JGMM}), \textit{sliced inverse
regression} (SIR) \citep{Li91} and \textit{multivariate relevance vector
machine} (RVM) \citep{thayananthan2006multivariate}. SIR is used with one (SIR-1) or two
(SIR-2) principal axes for dimensionality reduction, 20 slices
(the number of slices is known to have very little influence on
the results), and polynomial regression of order three (higher
orders did not show significant improvements in our experiments). SIR quantizes the
low-dimensional data $\Xvect$ into {\it slices} or clusters which
in turn induces a quantization of the $\Yvect$-space. Each
$\Yvect$-slice (all points $\yvect_n$ that map to the same
$\Xvect$-slice) is then replaced with its mean and PCA is carried
out on these means. The resulting dimensionality reduction is then
informed by $\Xvect$ values through the preliminary slicing. RVM \citep{thayananthan2006multivariate} may be view as a multivariate probabilistic formulation of \textit{support vector regression} \citep{smola2004tutorial}.
As all kernel methods, it critically depends on the choice of a kernel function. Using the authors'
freely available code\footnote{http://www.mvrvm.com/Multivariate\_Relevance\_Vector\ },
we ran preliminary tests to determine an optimal kernel choice for each
dataset considered. We tested 14 kernel types with 10 different scales ranging from 1 to
30, hence, 140 kernels for each dataset in total.

\subsection{High-dimensional Function Inversion}
\label{subsec:synth}
In this Section, we evaluate the ability of the different regression methods to learn a low-to-high dimensional function $\fvect$ from noisy training examples in order to inverse it. We consider a situation where some components $\wvect$ of the function's support are hidden during training, and where given a new image $\yvect=\fvect(\tvect,\wvect)$, $\tvect$ is to be recovered. In this Section, MLE and hGLLiM were constrained with equal and isotropic covariance
matrices $\{\Sigmamat_k\}_{k=1}^K$ as it showed to yield the best results with the synthetic functions considered.
For RVM, the kernel leading to
the best results out of $140$ tested kernels was the \textit{linear spline kernel}
\citep{vapnik1997support} with a scale parameter of 8, which is thus used in this Section.

The three vector-valued function families used for testing are of the form $\fvect:\mathbb{R}^2\rightarrow\mathbb{R}^D$,
$\gvect:\mathbb{R}^2\rightarrow\mathbb{R}^D$ and $\hvect:\mathbb{R}^3\rightarrow\mathbb{R}^D$. The three functions depend on $\xvect=[t;\wvect]$ which has an observed 1-dimensional part 
$t$, and an unobserved 1 or 2-dimensional part $\wvect$. Using the decomposition $\fvect=(f_1 \hdots f_d \hdots f_D)\tp$ for each function, each component is defined by:
\begin{eqnarray*}
  f_d(t,w_1)&=&\alpha_d\cos(\eta_d \; t/10+\phi_d)+\gamma_dw_1^3\\
  g_d(t,w_1)&=&\alpha_d\cos(\eta_d \; t/10+\beta_dw_1+\phi_d)\\
  h_d(t,w_1,w_2)&=&\alpha_d\cos(\eta_d \; t/10+\beta_dw_1+\phi_d)+\gamma_dw_2^3
\end{eqnarray*}
where $\xivect=\{\alpha_d,\eta_d,\phi_d,\beta_d,\gamma_d\}_{d=1}^D$ are
scalars in respectively $[0,2]$,
$[0,4\pi]$, $[0,2\pi]$, $[0,\pi]$ and $[0,2]$. This choice allows
to generate a wide range of high-dimensional functions with different properties, {\it e.g.}, monotonicity,
periodicity or sharpness. In particular,
the generated functions are chosen to be rather challenging for
the piecewise affine assumption made in hybrid GLLiM.

One hundred functions of each of these three types were
generated, each time using different values for $\xivect$ drawn uniformly at random. For each function, a set of
$N$ training couples $\{(t_n,\yvect_n)\}_{n=1}^{N}$ and a set of
$N'$ test couples \black{$\{(t'_n,\yvect'_n)\}_{n=1}^{N'}$} were synthesized by
randomly drawing $t$ and $\wvect$ values and by adding some random isotropic Gaussian noise
$\evect$, \textit{e.g.}, $\yvect = \fvect(t,\wvect)+\evect$. Values of $t$ were drawn uniformly in $[0,10]$, while values of $\wvect$ were drawn uniformly in $[-1,1]$ for $\fvect$ and $\gvect$, and in $[-1,1]^2$ for $\hvect$.
Training couples were used to train the different regression algorithms tested.
The task was then to compute an estimate $\hat{t'}_n$ given a test observation $\yvect'_n = \fvect(t'_n,\wvect'_n)+\evect'_n$.

Table \ref{tab:functions_err} displays the average (Avg), standard
deviation (Std) and percentage of \textit{extreme values} (Ex) of
the absolute errors $|\hat{t}'_n-t'_n|$ obtained with the different
methods. For each generated function, we used an observation dimension $D=50$, an average signal to noise ratio\footnote{$\textrm{SNR} = 10 \log \black{(\norm{\yvect}^2/\norm{\evect}^2})$} (SNR)
of \black{6 \textit{decibels} (dB)}, $N=200$ training
points and $N'=200$ test points, totaling $20,000$ tests per function type. MLE, JGMM and hGLLiM where used with $K=5$ mixture components.
We define \textit{extreme values} (Ex) as those higher than the average error
that would be obtained by an algorithm returning random values of
$t$ from the training set. \black{Since training values are uniformly spread in an interval
in all considered experiments, this corresponds to one third of the interval's length, \textit{e.g.}, 10/3 for the synthetic functions. This measure will be repeatedly used throughout the experiments.}

\begin{table}
\caption{\label{tab:functions_err}50-D synthetic data: Average
(Avg), standard deviation (Std) and percentage of extreme values (Ex) of the absolute error obtained with
different methods.}
   \centering
   \begin{tabular}{|c|c|c|c|c|c|c|c|c|c|}
       \cline{2-10}
       \multicolumn{1}{r}{}& \multicolumn{3}{|c|}{$\fvect$} &  \multicolumn{3}{c|}{$\gvect$} &  \multicolumn{3}{c|}{$\hvect$} \\
       \hline
       Method              & Avg  & Std  & Ex               & Avg  & Std  & Ex               & Avg  & Std  & Ex \\
       \hline
       JGMM     & $1.78$ & $2.21$ & $19.5$ & $2.45$ & $2.76$ & $28.4$ & $2.26$ & $2.87$ & $22.4$ \\
       \hline
       SIR-1    & $1.28$ & $1.07$ & $5.92$ & $1.73$ & $1.39$ & $14.9$ & $1.64$ & $1.31$ & $13.0$ \\
       \hline
       SIR-2    & $0.60$ & $0.69$ & $1.02$ & $1.02$ & $1.02$ & $4.20$ & $1.03 $& $1.06$ & $4.91$ \\
       \hline
       RVM      & $0.59$ & $0.53$ & $0.30$ & $0.86$ & $0.68$ & $0.52$ & $0.93$ & $0.75$ & $1.00$ \\
       \hline
       MLE      & $0.36$ & $0.53$ & $0.50$ & $0.36$ & $0.34$ & $0.04$ & $0.61$ & $0.69$ & $0.99$ \\
       \hline
       hGLLiM-1 & $0.20$ & $0.24$ & $0.00$ & $0.25$ & $0.28$ & $0.01$ & $0.46$ & $0.48$ & $0.22$ \\
       \hline
       hGLLiM-2 & $0.23$ & $0.24$ & $0.00$ & $0.25$ & $0.25$ & $0.00$ & $0.36$ & $0.38$ & $0.04$ \\
       \hline
       hGLLiM-3 & $0.24$ & $0.24$ & $0.00$ & $0.26$ & $0.25$ & $0.00$ & $0.34$ & $0.34$ & $0.01$ \\
       \hline
       hGLLiM-4 & $0.23$ & $0.23$ & $0.01$ & $0.28$ & $0.27$ & $0.00$ & $0.35$ & $0.34$ & $0.01$ \\
       \hline
       {\boldmath hGLLiM-BIC} & {\boldmath $0.18$} & {\boldmath $0.21$} & {\boldmath $0.00$} & {\boldmath $0.24$} & {\boldmath $0.26$} & {\boldmath $0.00$} & {\boldmath $0.33$} & {\boldmath $0.35$} & {\boldmath $0.06$} \\
       \hline
   \end{tabular}
\end{table}

\black{As showed in Table \ref{tab:functions_err}, all the hGLLiM-$L_\textrm{w}$ models with $L_\textrm{w}>0$ significantly outperform the five other regression techniques for the three functions considered, demonstrating the effectiveness of the proposed partially-latent variable model. 
For each generated training set, the hGLLiM-BIC method minimized BIC for $0\le L_{\textrm{w}}\le10$, and used the corresponding model to perform the regression. As showed in \ref{tab:functions_err}, hGLLiM-BIC outperformed all the other methods. In practice, hGLLiM-BIC did not use the same $L_{\textrm{w}}$ value for all functions of a given type. Rather, it was able to automatically select a value according to the importance of the latent components in the generated function. The second best method is MLE, \textit{i.e.}, hGLLiM-0. The relative decrease of average error between MLE and hGLLiM-BIC is of respectively $50\%$ for function $\fvect$, $33\%$ for function $\gvect$ and $46\%$ for function $\hvect$. This is a significant improvement, since errors are averaged over $20,000$ tests. Moreover, the addition of latent components in hGLLiM reduced the percentage of extreme errors. Interestingly, BIC selected the \textit{``expected'' latent dimension} $L^*_{\textrm{w}}$ for $72\%$ of the 300 generated functions, \textit{i.e.}, $L^*_{\textrm{w}}=1$ for $\fvect$ and $\gvect$ and $L^*_{\textrm{w}}=2$ for $\hvect$.}

\begin{figure}[t!]
 \centering
 \includegraphics[width=0.70\linewidth,clip=,keepaspectratio]{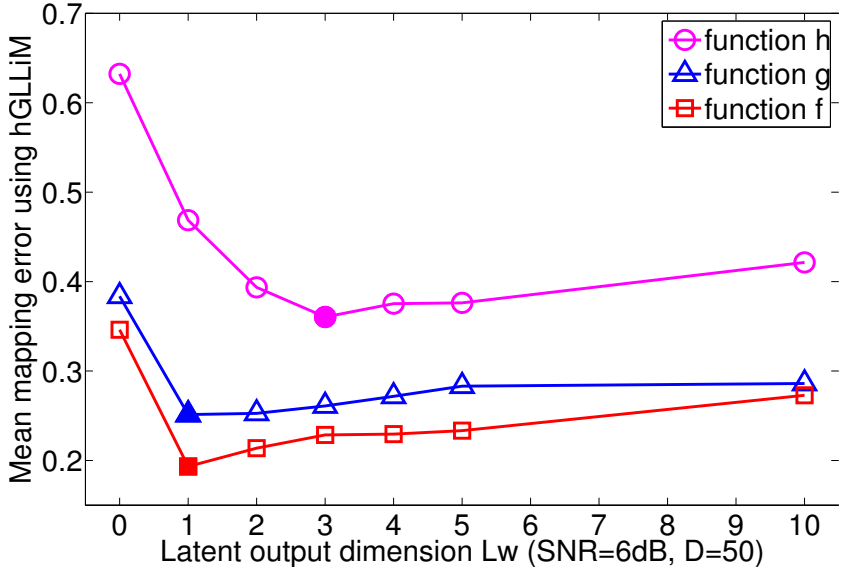}
 \caption{\label{fig:Lw_err1}Influence of the parameter $L_{\textrm{w}}$ of hGLLiM on the mean mapping error of synthetic functions $\fvect$, $\gvect$ and $\hvect$. The minimum of each plot is showed with a filled marker. Each point corresponds to an average error over $10,000$ tests on $50$ distinct functions.}
\end{figure}
Fig.~\ref{fig:Lw_err1} shows the results obtained using hGLLiM-$L_{\textrm{w}}$ and different values of $L_{\textrm{w}}$ for functions $\fvect$, $\gvect$ and $\hvect$.
For $\fvect$ and $\gvect$, the lowest average error is obtained using $L_\textrm{w}=1$. This is expected since $L^*_\textrm{w}=1$ for $\fvect$ and $\gvect$. \black{However, an interesting observation is made for function $\hvect$. Although $L^*_\textrm{w}=2$ for $\hvect$, even slightly lower average errors are obtained using $L_\textrm{w}=3$. While using the expected latent dimension $L_\textrm{w}=L^*_\textrm{w}$ always reduces the mean error with respect to
$L_\textrm{w}=0$ (MLE), the error may be farther reduced by selecting a latent dimension slightly larger that the expected one.
This suggests that the actual non linear latent
effects on the observations 
could be modeled  more accurately by choosing a latent dimension that is higher than
the  dimension expected intuitively.}

\begin{figure}[t!]
 \centering
 \includegraphics[width=0.70\linewidth,clip=,keepaspectratio]{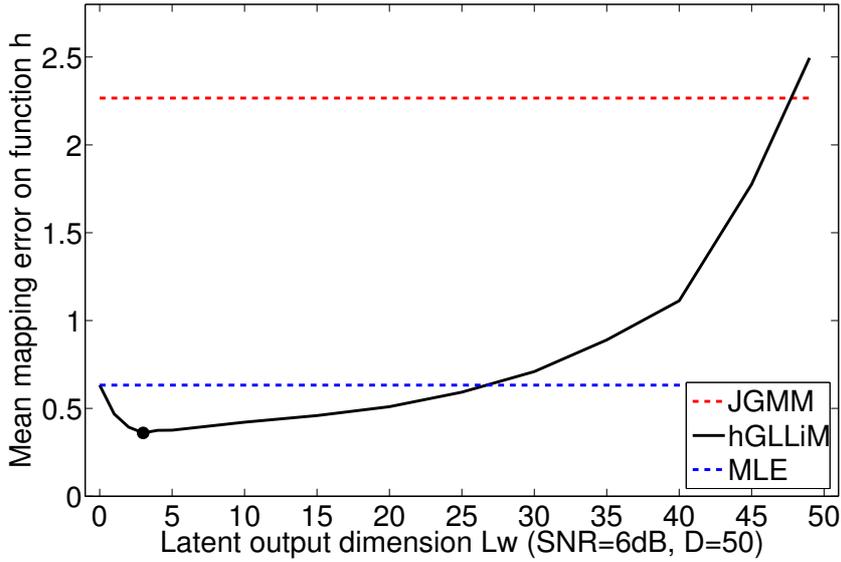}
 \caption{\label{fig:Lw_err2}Influence of the parameter $L_{\textrm{w}}$ of hGLLiM on the mean mapping error of synthetic functions $\hvect$. The minimum is showed with a filled marker. Mean mapping errors obtained with MLE and JGMM on the same data are also showed for comparison. Each point corresponds to an average error over $10,000$ tests on $50$ distinct functions.}
\end{figure}
Fig.~\ref{fig:Lw_err2} illustrates how hGLLiM provides a whole range of alternative models \textit{in between} MLE and JGMM, as explained in Section~\ref{sec:plvm}.
 Values of $L_\textrm{w}$ in the range $1\dots 20$ improve results
upon MLE which does not model unobserved variables. As
$L_\textrm{w}$ increases beyond $L^*_{\textrm{w}}$ the number of
parameters to estimate becomes larger and larger and the model
becomes less and less constrained until becoming equivalent to
JGMM with equal unconstrained covariances (see Section
\ref{sec:plvm} and Appendix
\ref{app:JGMM}).

\begin{figure}[t!]
 \centering
 \includegraphics[width=0.70\linewidth,clip=,keepaspectratio]{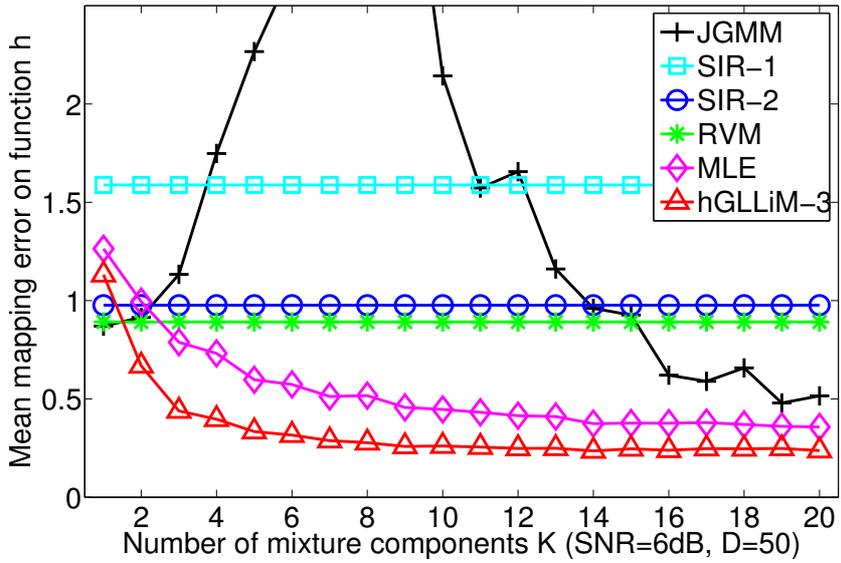}
 \caption{Influence of the number of initial mixture components $K$ in MLE, JGMM and hGLLiM-3 on the mean mapping error of synthetic function $\hvect$. Errors obtained with SIR-1, SIR-2 and RVM on the same data are also showed for comparison. Each point corresponds to an average error over $10,000$ tests on $50$ distinct functions.}
\end{figure}
\begin{figure}[t!]
 \centering
 \includegraphics[width=0.70\linewidth,clip=,keepaspectratio]{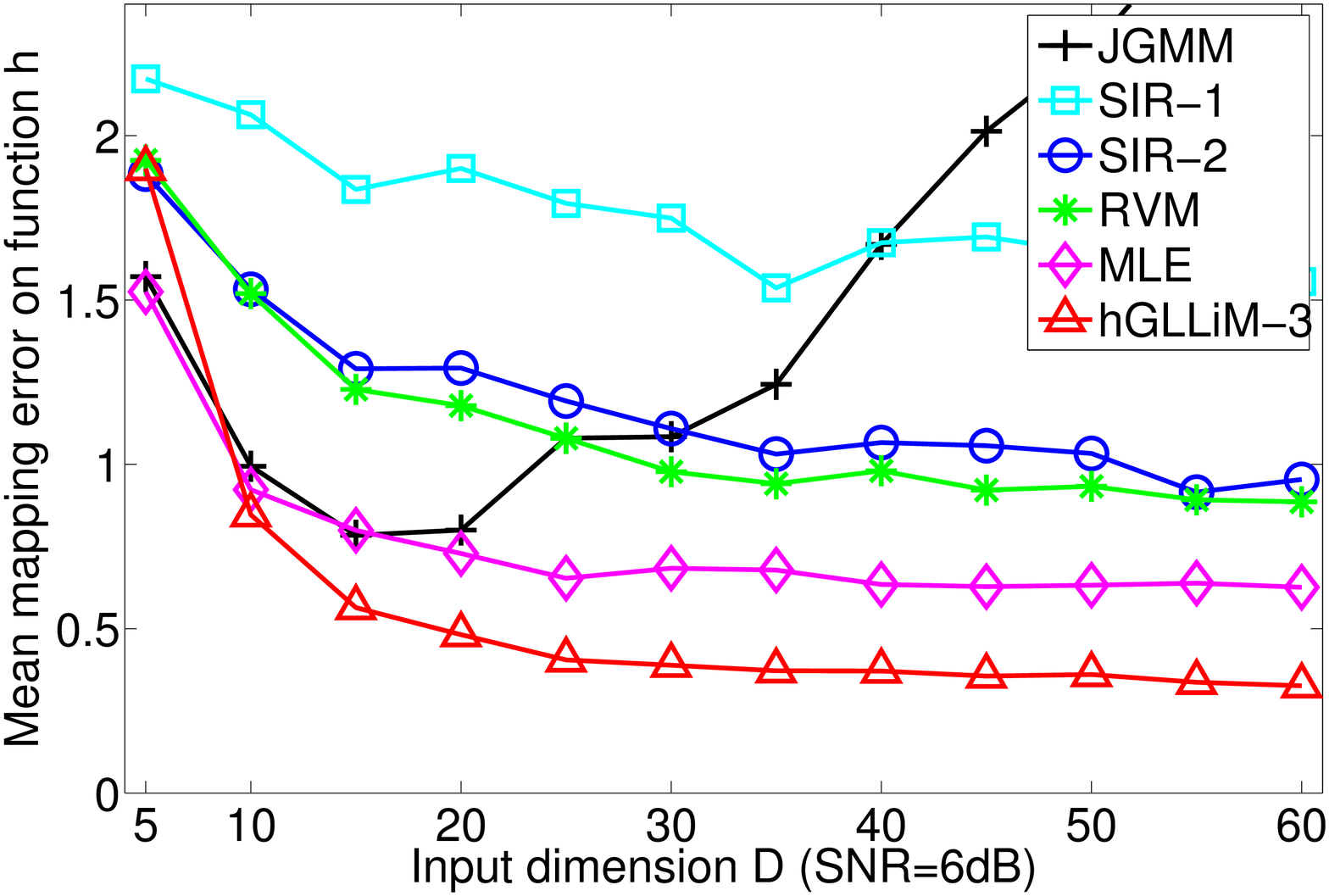}
 \caption{\label{fig:D_err}Influence of $D$ on the mean mapping error of synthetic functions $\hvect$ using different methods. Each point corresponds to an average error over $10,000$ tests on $50$ distinct functions.}
\end{figure}
\begin{figure}[t!]
 \centering
 \includegraphics[width=0.70\linewidth,clip=,keepaspectratio]{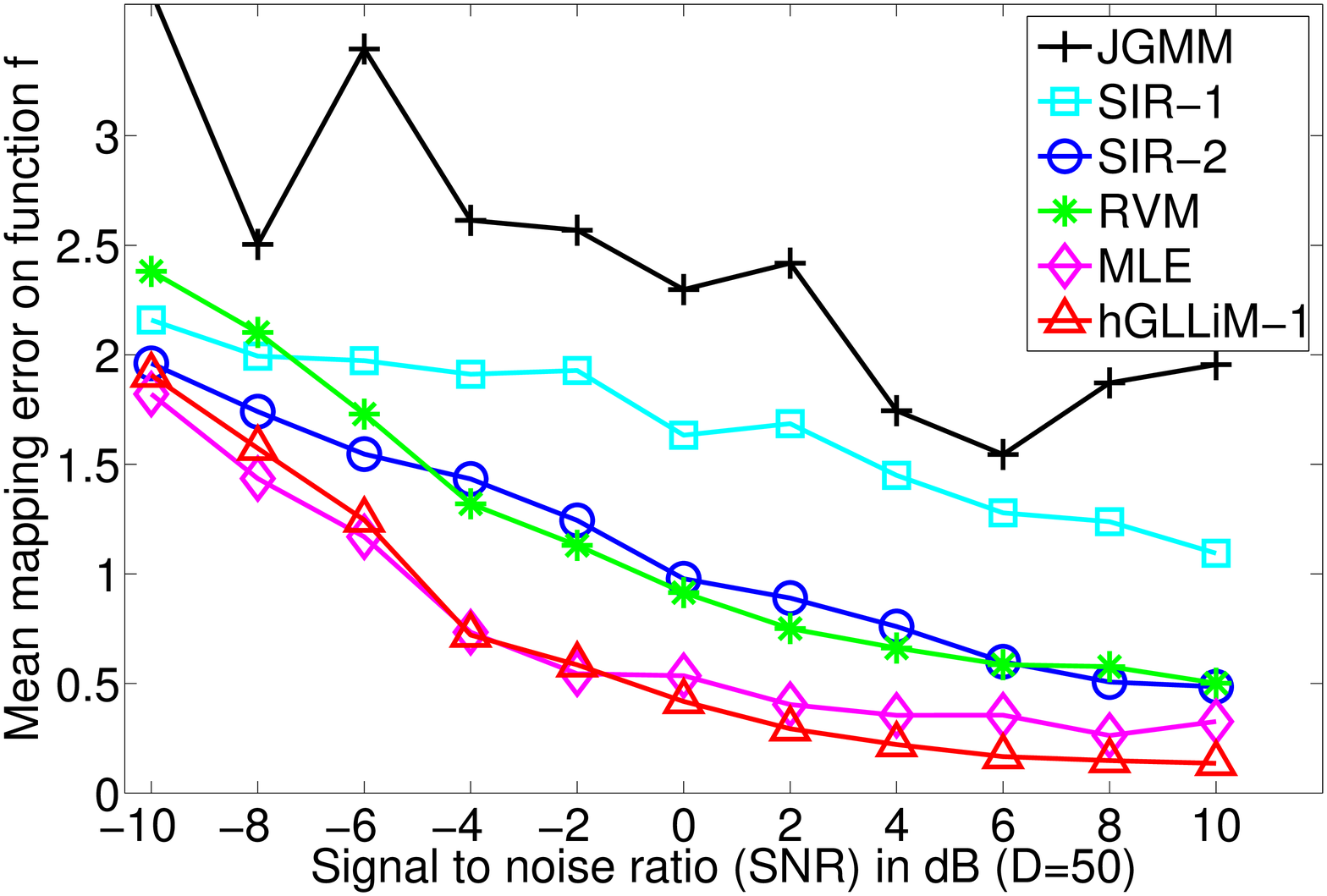}
 \caption{\label{fig:SNR_err}Influence of the signal-to-noise ration (SNR) on the mean mapping error of synthetic functions $\fvect$ using different methods. Each point corresponds to an average error over $10,000$ tests on $50$ distinct functions.}
\end{figure}
\black{Extensive experiments showed that obtained errors generally decrease when $K$
increases. However too high values of $K$ lead to degenerate
covariance matrices in classes where there are too few samples.
Such classes are simply removed along the execution of the
algorithms, thus reducing $K$. This is well illustrated in Fig.~\ref{fig:Lw_err2}: results obtained with hGLLiM do not significantly change for initial values of $K$ larger than 9 in that case. Similarly, although $K$ is manually set to a fixed value in the remaining experiments, further tests showed that higher values of $K$ always yielded either similar or better results, at the cost of more computational time. Fig.~\ref{fig:Lw_err2} also shows that the error made by JGMM severely increases with $K$ for $K<10$, and then decreases to become around $40\%$ larger than MLE. This is in fact an overfitting effect due to the very large numbers of parameters in JGMM. Indeed, the JGMM error with $K=20$ turned out to increase by more than $100\%$ when decreasing the SNR from 6dB to 3dB, while it increased by less than $30\%$ using all the other methods.}

Finally, Figures \ref{fig:D_err} and \ref{fig:SNR_err} show the influence
of the observation space dimension $D$ and the SNR on the mean mapping error using various methods.
While for low values of $D$ the 6 methods yield similar results,
the hybrid GLLiM approach significantly outperforms all of them in higher dimension (Average error $45\%$ lower than with MLE for all $D>30$).
Similarly, apart from JGMM which is very prone to overfitting due
to its large number of parameters when $D$ is high, all techniques
perform similarly under extreme noise level (SNR $=-10$ dB\black, where \textit{dB} means \textit{decibels}) while
hybrid GLLiM decreases the error up to $60\%$ compared to MLE for positive SNRs. 

\subsection{Robustly Retrieving Either Pose Or Light From Face Images}
\label{subsec:pose}
\begin{figure}[h]
  \centering
  \includegraphics[width=0.75\linewidth,clip=,keepaspectratio]{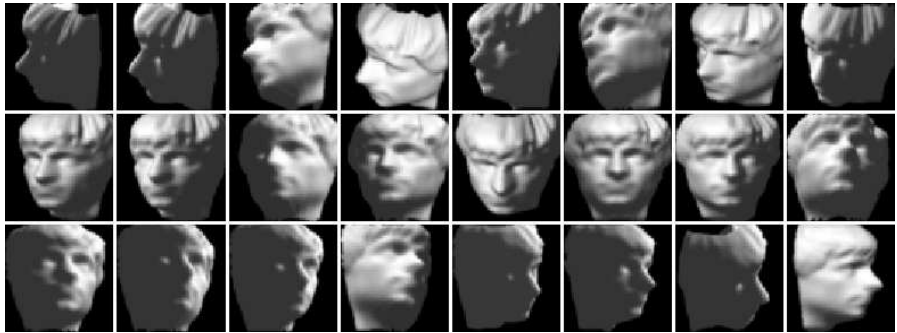}
  \caption{\label{fig:ex_face}Example of face images from the Stanford's face dataset.}
\end{figure}
We now test the different regression methods 
on the \textit{face
dataset}\footnote{http://isomap.stanford.edu/datasets.html} which
consists of 697 images (of size $64\times64$ pixels) of a 3D model of a head whose
pose is parameterized by a left-right \textit{pan} angle ranging from
$-75^\circ$ to $+75^\circ$ and an up-down \textit{tilt} angle ranging from
$-10^\circ$ to $+10^\circ$. Example of such images are given in Figure~\ref{fig:ex_face}.
The image of a face depends on the (pan,tilt) angles as well as on lighting that is absolutely necessary for rendering.
The latter is simulated with one parameter
taking integer values between $105$ and $255$. Images were down-sampled\footnote{\black{We kept one every 4 pixels horizontally and vertically.}} to $16\times16$ and stacked
into $D=256$ dimensional vectors. In the tasks considered, the algorithms were trained using
a random subset of $N=597$ images, and tested with the remaining $N'=100$
images. We repeated this train-then-test process 50 times for each task ($5,000$ tests per task in total).
We used $K=10$ for MLE, hGLLiM and JGMM (see discussion on $K$ in Section \ref{subsec:synth}). Again, hGLLiM and MLE were constrained with equal and isotropic covariance
matrices $\{\Sigmamat_k\}_{k=1}^K$ as it showed to yield the best results. Regarding RVM, as done previously,
the best out of 140 kernels was used, {\it i.e.}, linear spline with scale 20.

We consider two tasks where the target variable is only partially annotated. Firstly, the methods are used to learn the image-to-pose mapping using pairs of image-pose observations for training while the lighting is unobserved, \textit{i.e.},
\textit{light-invariant face pose
estimation}. Secondly, the methods are used to learn the image-to-lighting mapping using pairs of image-light observations for training while the pose is unobserved, \textit{i.e.},
\textit{pose-invariant light-direction
estimation}. Table~\ref{tab:face_err} shows results
obtained with the different methods. We show results obtained with hGLLiM-$L_\textrm{w}^*$, hGLLiM-$L_\textrm{w}^\dagger$ and hGLLiM-BIC. $L_\textrm{w}^*$ denotes the expected latent dimension, and
$L_\textrm{w}^\dagger$ is the latent dimension which empirically showed the best results, when varying $L_\textrm{w}$ between $0$ and $30$ (larger values showed to systematically increase the error). For each training set, the hGLLiM-BIC method minimized BIC for $0\le L_{\textrm{w}}\le30$, and used the corresponding model to perform the regression.
For light-invariant face pose estimation the expected latent dimension is $L_\textrm{w}^*=1$, and we obtained the best results with $L_\textrm{w}^\dagger=13$ (values in $[6,20]$ yielded similar errors).
For pose-invariant light-direction estimation the expected latent dimension is $L_\textrm{w}^*=2$, and we obtained the best results with $L_\textrm{w}^\dagger=19$ (values in $[11,20]$ yielded similar errors). As in Section \ref{subsec:synth}, we observe that while the expected latent dimension improves upon $L_\textrm{w}=0$, the error may be farther reduced by selecting a latent dimension larger that the true one. Overall, hGLLiM-$L_\textrm{w}^\dagger$ achieved a
$20\%$ to  $60\%$ improvement with respect to MLE on this standard dataset. 
\black{This time, hGLLiM-BIC performed worse than hGLLiM-$L_\textrm{w}^*$ and hGLLiM-$L_\textrm{w}^\dagger$ and performed only slightly better than MLE. The expected  latent dimension 1 was estimated in $70\%$ of the case for the face pose estimation task, but BIC found a latent dimension of 0 or 1 instead of 2 for the light-direction estimation task.}

\begin{table}
   \caption{\label{tab:face_err} Face dataset: Average
(Avg), standard deviation (Std) and percentage of extreme values
(Ex) of absolute pan and tilt angular errors and light errors
obtained with different methods. Superscript ${}^*$ stands for
$L_w$ set to its true value $L_w^*$ while ${}^\dagger$ stands for
$L_w$ set to the best found dimension in terms of empirical
error.} \centering
   \begin{tabular}{|c|c|c|c|c|c|c|c|c|c|c|c|}
       \cline{2-7} \cline{10-12}
       \multicolumn{1}{r}{}& \multicolumn{3}{|c|}{Pan error $({}^{\circ})$}&  \multicolumn{3}{c|}{Tilt error $({}^{\circ})$} & \multicolumn{1}{r}{} & &  \multicolumn{3}{|c|}{Light error}  \\
       \cline{1-7} \cline{9-12}
       Method & Avg& Std & Ex     & Avg            & Std           & Ex  & & Method & Avg            & Std           & Ex   \\
       \cline{1-7} \cline{9-12}
       JGMM   & $13.2$ & $26.6$ & $8.2$ & $2.32$ & $3.01$ & $7.0$ & & JGMM  & $18.2$ & $21.0$ & $6.7$ \\
       \cline{1-7} \cline{9-12}
       SIR-1  & $16.0$ & $11.3$ & $1.4$ & $2.64$ & $2.06$ & $4.9$ & & SIR-1 & $15.2$ & $13.2$ & $3.2$\\
       \cline{1-7} \cline{9-12}
       SIR-2  & $10.6$ & $9.73$ & $0.4$ & $1.81$ & $1.66$ & $1.9$ & & SIR-2 & $13.6$ & $13.2$ & $2.8$\\
       \cline{1-7} \cline{9-12}
       RVM    & $14.0$ & $12.2$ & $1.9$ & $2.63$ & $2.13$ & $5.8$ & & RVM & $18.7$ & $15.7$ & $4.82$\\
       \cline{1-7} \cline{9-12}
       MLE    & $6.01$ & $5.35$ & $0.0$ & $1.84$ & $1.64$ & $1.8$ & & MLE & $10.9$ & $8.84$ & $0.2$\\
       \cline{1-7} \cline{9-12}
       hGLLiM-$1^*$  & $3.80$ & $4.33$ & $0.0$ & $1.58$ & $1.46$ & $1.0$ & & hGLLiM-$2^*$ & $10.1$ & $8.84$ & $0.2$\\
       \cline{1-7} \cline{9-12}
       \textbf{hGLLiM-}{\boldmath $13^{\dagger}$} & {\boldmath $2.65$} & {\boldmath $2.39$} & {\boldmath  $0.0$}  & {\boldmath $1.19$} & {\boldmath $1.11$} & {\boldmath  $0.2$} & &
       \textbf{hGLLiM-}{\boldmath $19^{\dagger}$} & {\boldmath $8.71$} & {\boldmath $7.54$} & {\boldmath  $0.0$} \\
       \cline{1-7} \cline{9-12}
       hGLLiM-BIC & $4.11$ & $4.66$ & $0.0$ & $1.58$ & $1.47$ & $1.0$ & &  hGLLiM-BIC & $10.3$ & $8.66$ & $0.2$ \\
       \cline{1-7} \cline{9-12}
   \end{tabular}
\end{table}

\begin{figure}[t!]
   \center
   \begin{tabular}{|c|l|c|c|}
    \hline
    Input & hGLLiM-1   & Reconstructions for different values of $w$ ($L_{\textrm{w}}=1$)                       & Recons. \\
    image & estimates  & \includegraphics[height = 0.0165\linewidth,clip=,keepaspectratio]{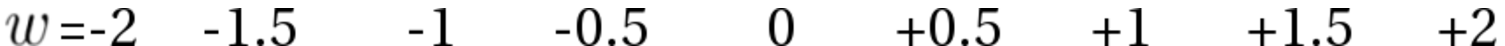} & MLE\\
    \hline
    \multirow{3}{*}{\includegraphics[height = 0.065\linewidth,clip=,keepaspectratio]{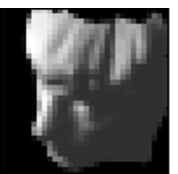}}   & $t_1=-41^\circ$&
    \multirow{3}{*}{\includegraphics[height = 0.065\linewidth,clip=,keepaspectratio]{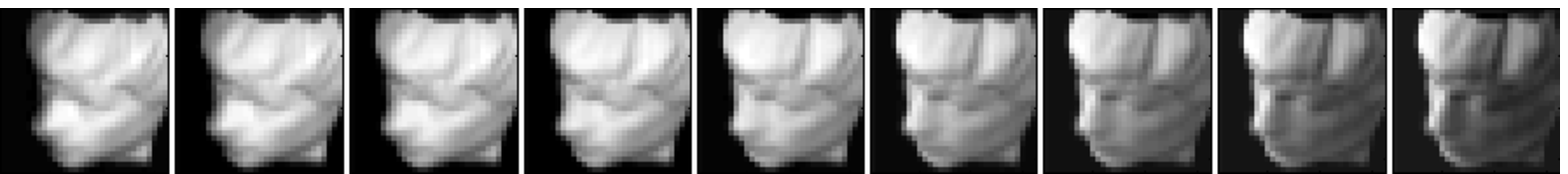}} &
    \multirow{3}{*}{\includegraphics[width = 0.065\linewidth,clip=,keepaspectratio]{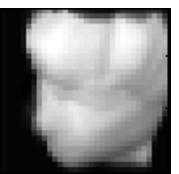}} \\
    & $t_2=8.7^\circ$ & & \\
    & $w=1.73$ & & \\
    \hline
    \multirow{3}{*}{\includegraphics[height = 0.065\linewidth,clip=,keepaspectratio]{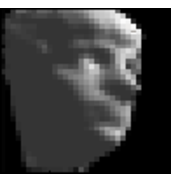}}   & $t_1=55^\circ$&
    \multirow{3}{*}{\includegraphics[height = 0.065\linewidth,clip=,keepaspectratio]{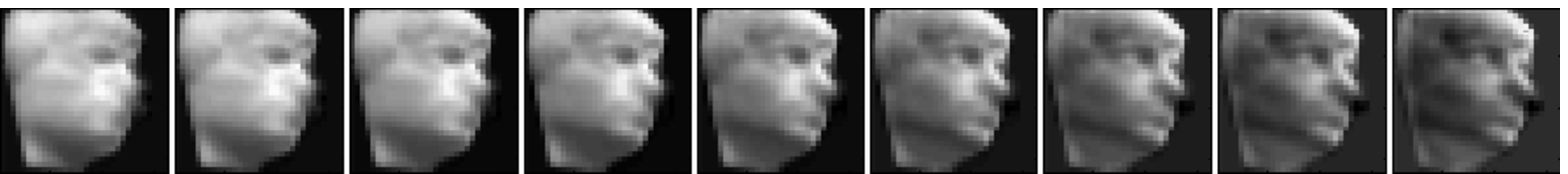}} &
    \multirow{3}{*}{\includegraphics[width = 0.065\linewidth,clip=,keepaspectratio]{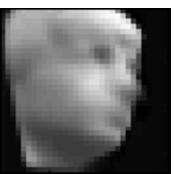}} \\
    & $t_2=-5.4^\circ$ & & \\
    & $w=0.28$ & & \\
    \hline
    \multirow{3}{*}{\includegraphics[height = 0.065\linewidth,clip=,keepaspectratio]{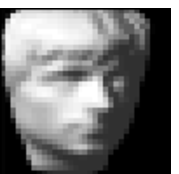}}   & $t_1=-9.8^\circ$&
    \multirow{3}{*}{\includegraphics[height = 0.065\linewidth,clip=,keepaspectratio]{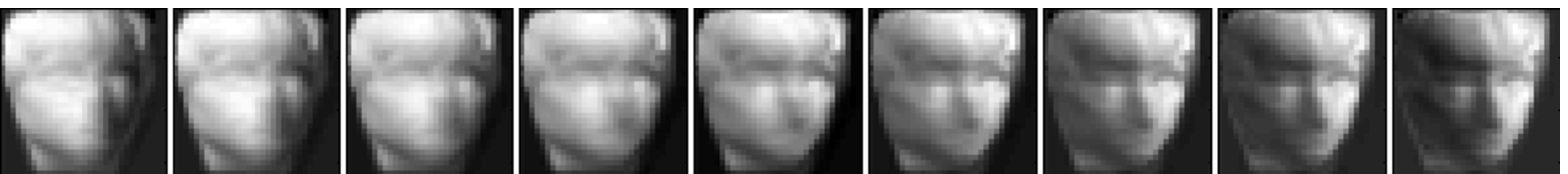}} &
    \multirow{3}{*}{\includegraphics[width = 0.065\linewidth,clip=,keepaspectratio]{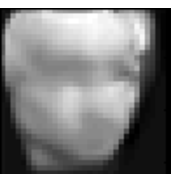}} \\
    & $t_2=4.3^\circ$ & & \\
    & $w=-1.47$ & & \\
    \hline
    \multirow{3}{*}{\includegraphics[height = 0.065\linewidth,clip=,keepaspectratio]{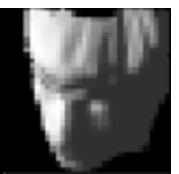}}   & $t_1=-24^\circ$&
    \multirow{3}{*}{\includegraphics[height = 0.065\linewidth,clip=,keepaspectratio]{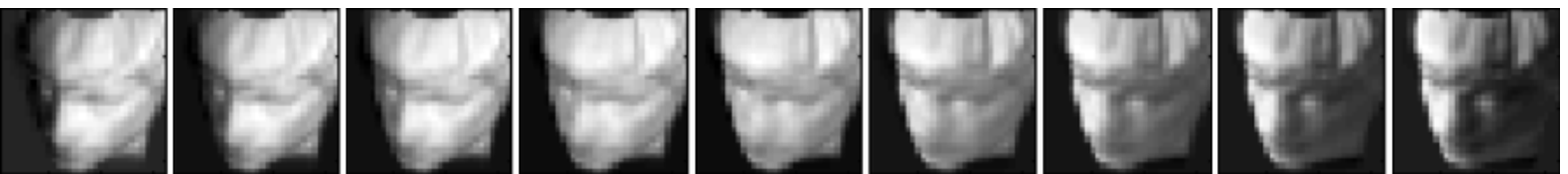}} &
    \multirow{3}{*}{\includegraphics[width = 0.065\linewidth,clip=,keepaspectratio]{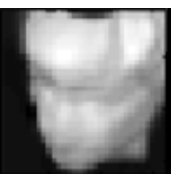}} \\
    & $t_2=8.2^\circ$ & & \\
    & $w=1.32$ & & \\
    \hline
    \multicolumn{1}{c}{(a)} & \multicolumn{1}{c}{(b)} & \multicolumn{1}{c}{(c)} & \multicolumn{1}{c}{(d)}\\
   \end{tabular}
\caption{\label{fig:face} Recovering the pose of a face ($t_1$=pan angle, $t_2$=tilt angle) with lighting being modeled by the latent variable $W$. (a)~The input image. (b)~The pose and lighting estimates using hybrid GLLiM. (c) Reconstructed images using the estimated pose parameters and different values for $w$. (d) Reconstructed images using the pose parameters estimated using MLE.}
\end{figure}

Another experiment was run to verify whether the latent variable values recovered with our method were meaningful. Once a set of model parameters $\widetilde{\thetavect}$ were estimated using hGLLiM-1 and with a training set of 597 pose-to-image associations, a different test image $\yvect'$ was selected at random and was used to recover $\hat{\tvect'}\in\mathbb{R}^2$ and $\hat{w'}\in\mathbb{R}$ based on the
forward regression function (\ref{eq:JGMM_inverse_exp}), \textit{i.e.},
$\hat{\xvect'}=[\hat{\tvect'};\hat{w'}]=\mathbb{E}[\Xvect|\yvect';\widetilde{\thetavect}]$
(see Section \ref{sec:gllim}). An image was then reconstructed using the
inverse regression function (\ref{eq:JGMM_forward_exp}), \textit{i.e.},
$\hat{\yvect'}=\mathbb{E}[\Yvect|[\hat{\tvect'};w];\widetilde{\thetavect}]$,
while varying the value of $w$ in order to visually observe its
influence on the reconstructed image. Results obtained for different test images are
displayed in Fig.~\ref{fig:face}. These results show that the
latent variable $\Wvect$ of hybrid GLLiM does capture lighting effects,
whereas an explicit lighting parameterization was not present in the
training set. For comparison, we show images obtained after
projection and reconstruction when MLE
(or $L_{\textrm{w}}=0$) is used instead. As it may be observed, the image reconstructed with MLE looks like
a blurred average over all possible lightings, while hybrid GLLiM allows a
much more accurate image reconstruction process. This is because
hybrid GLLiM encodes images with $3$ rather than $2$ variables, one of which being latent and estimated in an unsupervised way.

\subsection{Retrieval of Mars surface physical properties from hyperspectral images}
\label{subsec:planet}
Visible and near infrared imaging
spectroscopy is a key remote sensing technique used to study and
monitor planets.  \textcolor{black}{It records the visible and
infrared light reflected from the planet in a given wavelength
range and produces cubes of data where each observed surface location is associated with a spectrum.} Physical properties of
the planets' surface, such as chemical composition, granularity,
texture, etc, are some of the most important parameters that
characterize the morphology of spectra. In the case of Mars,
radiative transfer models have been developed to numerically evaluate the link between these parameters and observable
spectra. Such models allow to simulate spectra from a given set of
parameter values, \textit{e.g.}, \citep{Doute2007}. In practice, the goal is to
scan the Mars ground from an orbit in order to observe gas and dust
in the atmosphere and look for signs of specific materials such as
silicates, carbonates and ice at the surface. We are thus
interested in solving the associate inverse problem which is to
deduce physical parameter values from the observed spectra.
Since this inverse problem cannot generally be solved analytically, the use of
optimization or statistical methods has been investigated,
{\it e.g.} \citep{bernard2009retrieval}. In
particular, training approaches have been considered with the
advantage that, once a relationship between parameters and spectra
has been established trough training, the learn relationship can be used for very
large datasets and for all new images having the same physical
model.

Within this category of methods,  we investigate the potential of the proposed
hybrid GLLiM model using a dataset of hyperspectral images
collected from the imaging spectrometer OMEGA instrument
\citep{Bibring2004} onboard of the Mars express spacecraft. To this
end a database of synthetic spectra with their associated
parameter values were generated using a radiative transfer
model. This database is composed of 15,407 spectra associated with five
real  parameter values, namely, proportion of water
ice, proportion of CO$_2$ ice, proportion of
dust, grain size of water ice, and grain size of CO$_2$ ice. Each
spectrum is made of 184 wavelenghts. \textcolor{black}{The hybrid GLLiM
method can be used, first to learn as {\it inverse} regression between
parameters and spectra from the database, and second to estimate the
corresponding parameters for each new spectrum using the learned
relationship. Since no ground truth is available for Mars, the
synthetic database will also serve as a first test set to
evaluate the accuracy of the predicted parameter values. In order to fully illustrate the
potential of hybrid GLLiM, we deliberately ignore two of the parameters in
the database and consider them as latent variables.  We chose to
ignore the proportion of water ice and the grain size of CO$_2$
ice. These two parameters appear in some previous study
\citep{bernard2009retrieval} to be sensitive to the same
wavelengths than the proportion of dust and are suspected to mix
with the other parameters in the synthetic transfer model so that
they are harder to estimate.
We observed that
using them in the inversion tend to degrade the estimation of the
other three parameters, which are of particular interest, namely
proportion of CO$_2$ ice, proportion of dust and grain size of
water ice}. Therefore, we excluded the proportion of water ice and
the grain size of CO$_2$ ice, treated them as latent variables,
and did the regression with the three remaining parameters.

Hybrid GLLiM was then compared to
JGMM, SIR-1, SIR-2, RVM and MLE. An objective evaluation was done
by cross validation. We selected 10,000 training couples at random
from the training set, tested on the 5,407 remaining
spectra, and repeated this 20 times. 
\black{For all algorithms, training data were normalized to have 0 mean and unit variance using scaling and translating factors. These factors were then used on test data and estimated output to obtain final estimates. This technique showed to noticeably improve results of all methods. We used $K=50$ for MLE, HGLLiM and JGMM. MLE and JGMM were constrained with equal, diagonal covariance matrices as it showed to yield the best results.
For each training set, the hGLLiM-BIC method minimized BIC for $0\le L_{\textrm{w}}\le20$, and used the corresponding model to perform the regression.} As regards RVM, the best out
of 140 kernels was used. A third degree polynomial kernel with scale
6 showed the best results using cross-validation on a subset of
the database. As a quality measure of the
estimated parameters, we computed normalized root mean squared
errors (NRMSE\footnote{$\mbox{NMRSE }=\sqrt{\frac{\sum_{m=1}^M
(\hat{t}_m - t_m)^2}{\sum_{m=1}^M (t_m - \overline{t})^2}}$ with
$\overline{t} = M^{-1} \sum_{m=1}^M t_m$.}). The NRMSE quantifies
the difference between the estimated and real parameter values.
This measure is normalized enabling direct comparison between the
parameters which are of very different range. The closer NRMSE is
to zero the more accurate are the predicted values. Table
\ref{tab:planeto_err} shows obtained NRMSE for the three parameters
considered. \black{The expected latent variable dimension is $L^{*}_{\textrm{w}}=2$, and accordingly, the empirically best dimension for hGLLiM was $L^{\dagger}_{\textrm{w}}=2$. hGLLiM-2 outperformed all the other methods on that task, with an error $36\%$ lower than the second best method RVM, closely followed by MLE. No significant difference was observed between hGLLiM-2 and hGLLiM-3.
Note that due to the computation of the $D\times D$ kernel matrix, the computational and memory
costs of RVM for training were about 10 times higher than
those of hGLLiM, using Matlab implementations. Interestingly, BIC performed very well on these large training sets ($N=10,000$) as it correctly selected $L_{\textrm{w}}=2$ for the 20 considered training sets, yielding identical results as the best method hGLLiM-2.}

\begin{table}
\caption{\label{tab:planeto_err} Normalized root mean squared error (NRMSE) for Mars surface physical properties recovered from
hyperspectral images, using synthetic data and different methods.}
   \centering
   \begin{tabular}{|c|c|c|c|}
       \hline
       Method & Proportion of CO2 ice & Proportion of dust & Grain size of water ice \\
       \hline
       JGMM   & $0.83\pm1.61$ & $0.62\pm1.00$ & $0.79\pm1.09$  \\
       \hline
       SIR-1  & $1.27\pm2.09$ & $1.03\pm1.71$ & $0.70\pm0.94$  \\
       \hline
       SIR-2  & $0.96\pm1.72$ & $0.87\pm1.45$ & $0.63\pm0.88$  \\
       \hline
       RVM    & $0.52\pm0.99$ & $0.40\pm0.64$ & $0.48\pm0.64$ \\
       \hline
       MLE    & $0.54\pm1.00$ & $0.42\pm0.70$ & $0.61\pm0.92$ \\
       \hline
       hGLLiM-1 & $0.36\pm0.70$ & $0.28\pm0.49$ & $0.45\pm0.75$  \\
       \hline
       \textbf{hGLLiM-2}{\boldmath ${}^{*\dagger}$} &  {\boldmath $0.34\pm0.63$} & {\boldmath $0.25\pm0.44$} & {\boldmath $0.39\pm0.71$} \\
       \hline
       hGLLiM-3 &  $0.35\pm0.66$ & $0.25\pm0.44$ & $0.39\pm0.66$ \\
       \hline
       hGLLiM-4 & $0.38\pm0.71$ & $0.28\pm0.49$ & $0.38\pm0.65$ \\
       \hline
       hGLLiM-5 & $0.43\pm0.81$ & $0.32\pm0.56$ & $0.41\pm0.67$ \\
       \hline
       hGLLiM-20 & $0.51\pm0.94$ & $0.38\pm0.65$ & $0.47\pm0.71$ \\
       \hline
       \textbf{hGLLiM-BIC} &  {\boldmath $0.34\pm0.63$} & {\boldmath $0.25\pm0.44$} & {\boldmath $0.39\pm0.71$} \\
       \hline
\end{tabular}
\end{table}

Finally, we used an adequately selected subset of the synthetic
database,{\it e.g.}, \citep{bernard2009retrieval} to train the
algorithms, and test them on real data made of observed spectra.
In particular, we focus on a dataset of Mars South polar cap.
Since no ground truth is currently available for the physical
properties of Mars polar regions,  \black{we propose a qualitative
evaluation using  hGLLiM-2 and the three best performing methods,
among the tested ones, namely RVM, MLE and JGMM.
This evaluation is detailed in Section 2 of the Supplementary Materials.
hGLLiM-2 appears to matches satisfyingly expected
results from planetology.}

\section{Conclusion}
\label{sec:conclusion}



The main idea of the method proposed in this paper is to introduce the concept of partially-latent response augmentation in regression. Starting with the mixture of linear regressors family of techniques, we introduced the hybrid GLLiM model. The methodological implementation of the proposed model is investigated. We devised and described in detail an expectation-maximization inference procedure that can be viewed as a generalization of a number of existing probabilistic mapping techniques that span both regression and dimensionality reduction. The method is particularly well suited for estimating the parameters of high-dimensional to low-dimensional mapping problems, all in the presence of training data that contain both pertinent and irrelevant information for the problem at hand. The practical advantages of adding a latent component to the observed outputs is thoroughly tested with both simulated and real data and compared with a large number of probabilistic and deterministic regression methods. In the light of these experiments one may conclude that the proposed algorithm outperforms several existing techniques. This paves the road towards a deeper understanding of a wide range of applications for which training data, that capture the full complexity of natural phenomena, are merely available. The introduction of a latent component allows to capture data behaviors that cannot be easily modeled; in the same time it introduces some form of slack in the parameter inference procedure. 
\textcolor{black}{As regards the  automatic estimation of the latent component dimension, the
generative nature of our probabilistic model allows to treat this issue
as a model selection problem and to consider standard
information criteria, such as the Bayesian information criterion. This criterion showed very interesting results and good performance especially for large training data sets. However, it imposes to run a number of different models to select the best one and may therefore be computationally costly.}

\textcolor{black}{Further research could then include the investigation of adaptive ways to select the latent dimension or other criteria as mentioned by \citep{BouveyronCeleuxGirard2011}
 for estimating the intrinsic dimension in high
dimensional data. Another useful extension would be to take into account more complex dependencies between variables especially when  data correspond to images with some spatial structure. Also, similarly to \cite{ingrassia2012local}, more complex noise models
could be investigated via Student distributions ({\it e.g.} \cite{McLachlanPeel1998}) to allow for outliers accommodation and more robust estimation. Finally, it would be interesting to assess the behavior of our method in the presence of irrelevant regressors especially by comparison to other standard methods  which are not designed to handle such regressors. }


\appendix
\normalsize
\section{Link between joint GMM and GLLiM}
\label{app:JGMM}
\begin{proposition}
A GLLiM model on $\Xvect,\Yvect$ with unconstrained parameters $\thetavect = \{\cvect_k,\Gammamat_k,\pi_k,\Amat_k$,$\bvect_k,\Sigmamat_k\}_{k=1}^K$ is equivalent to a Gaussian mixture model on the joint variable $[\Xvect;\Yvect]$
with unconstrained parameters $\psivect=\{\mvect_k,\Vmat_k,\rho_k\}_{k=1}^K$, \textit{i.e.},
\begin{equation}
p(\Xvect=\xvect,\Yvect=\yvect;\thetavect)=\sum_{k=1}^K \rho_k \mathcal{N}([\xvect;\yvect];\mvect_k,\Vmat_k).
\end{equation}
The parameter vector $\thetavect$ can be expressed as a function of $\psivect$ by:
\begin{equation}
 \label{eq:GLMtoJGMM}
\begin{array}{rl}
 \pi_k =&\rho_k, \\
 \cvect_k =&\mvect^{\textrm{x}}_{k}, \\
 \Gammamat_k =&\Vmat_k^{\textrm{xx}}, \\
 \Amat_k =& \Vmat_k^{\textrm{xy}\top}(\Vmat_k^{\textrm{xx}})\inverse, \\
 \bvect_k =& \mvect^{\textrm{y}}_{k}-(\Vmat_k^{\textrm{xy}})\tp(\Vmat_k^{\textrm{xx}})\inverse\mvect^{\textrm{x}}_{k},\\
 \Sigmamat_k =& \Vmat_k^{\textrm{yy}}-(\Vmat_k^{\textrm{xy}})\tp(\Vmat_k^{\textrm{xx}})\inverse\Vmat_k^{\textrm{xy}},\\
 \textrm{where} \hspace{2mm} \mvect_k=&\left[
  \begin{array}{c}
   \mvect^{\textrm{x}}_{k}\\
   \mvect^{\textrm{y}}_{k}
  \end{array}
 \right] \\
 \textrm{and} \hspace{2mm} \Vmat_k=&\left[
  \begin{array}{lc}
    \Vmat_k^{\textrm{xx}}     & \Vmat_k^{\textrm{xy}}\\
    \Vmat_k^{\textrm{xy}\top} & \Vmat_k^{\textrm{yy}}
  \end{array}
 \right].
 \end{array}
\end{equation}
The parameter $\psivect$ can be expressed as a function of $\thetavect$  by:
\begin{equation}
 \label{eq:JGMMtoGLM}
 \begin{array}{rl}
 \rho_k=&\pi_k \\
 \mvect_k=&\left[
  \begin{array}{c}
   \cvect_k \\
   \Amat_k\cvect_k+\bvect_k
  \end{array}
 \right] \\
  \Vmat_k=&\left[
  \begin{array}{cc}
   \Gammamat_k & \Gammamat_k\Amat_k\tp\\
   \Amat_k\Gammamat_k & \Sigmamat_k+\Amat_k\Gammamat_k\Amat_k\tp
  \end{array}
 \right].
 \end{array}
\end{equation}
\end{proposition}
\color{black}
Note that this proposition was proved for $D=1$ in \citep{ingrassia2012local}, but not in the general case as proposed here.
\color{black}
\paragraph{Proof.}
(\ref{eq:GLMtoJGMM}) is obtained using (\ref{eq:JGMMtoGLM}) and
formulas for conditional multivariate Gaussian variables.
(\ref{eq:JGMMtoGLM}) is obtained from standard algebra by
identifying the joint distribution
$p(\Xvect,\Yvect|\Zvect;\thetavect)$ defined by
(\ref{eq:model_pYxz}) and (\ref{eq:model_pXZ}) with a multivariate
Gaussian distribution. To complete the proof, one need to prove
the following two statements:
\begin{itemize}
\item[{\bf (i)}] For any $\rho_k\in\mathbb{R}, \mvect_k\in\mathbb{R}^{D+L}$ and $\Vmat_k\in\mathcal{S}^{L+D}_+$,
there is a set of parameters $\cvect_k\in\mathbb{R}^L,\Gammamat_k\in\mathcal{S}^{L}_+,\pi_k\in\mathbb{R},\Amat_k\in\mathbb{R}^{D\times L},\bvect_k\in\mathbb{R}^D$ and $\Sigmamat_k\in\mathcal{S}^{D}_+$ such that (\ref{eq:GLMtoJGMM}) holds.
\item[{\bf (ii)}] Reciprocally, for any $\cvect_k\in\mathbb{R}^L,\Gammamat_k\in\mathcal{S}^{L}_+,\pi_k\in\mathbb{R},\Amat_k\in\mathbb{R}^{D\times L},\bvect_k\in\mathbb{R}^D,\Sigmamat_k\in\mathcal{S}^{D}_+$ there is a set of parameters $\rho_k\in\mathbb{R}, \mvect_k\in\mathbb{R}^{L+D}$ and $\Vmat_k\in\mathcal{S}^{D+L}_+$ such that (\ref{eq:JGMMtoGLM}) holds,
\end{itemize}
where $\mathcal{S}^{M}_+$ denotes the set of $M\times M$ symmetric positive definite matrices. We introduce the following lemma:
\begin{lemma}
\label{lemma:lemma1}
\label{lem:GLM_JGMM1} If
$$\Vmat=\left[
  \begin{array}{lc}
    \Vmat^{\textnormal{xx}}     & \Vmat^{\textnormal{xy}}\\
    \Vmat^{\textnormal{xy}\top} & \Vmat^{\textnormal{yy}}
  \end{array}
 \right]\in\mathcal{S}^{L+D}_+,$$ then
 $\Sigmamat=\Vmat^{\textnormal{yy}} - \Vmat^{\textnormal{xy}\top}\Vmat^{\textnormal{xx}-1}\Vmat^{\textnormal{xy}}\in\mathcal{S}^{D}_+$.
\end{lemma}
\paragraph{Proof.}
Since $\Vmat\in\mathcal{S}^{L+D}_+$ we have $\uvect\tp\Vmat\uvect>0$ for all non null $\uvect\in\mathbb{R}^{L+D*}$. Using the decomposition $\uvect=[\uvect^{\textrm{x}};\uvect^{\textrm{y}}]$ we obtain
$$
\uvect^{\textrm{x}\top}\Vmat^{\textrm{xx}}\uvect^{\textrm{x}}+
2\uvect^{\textrm{x}\top}\Vmat^{\textrm{xy}}\uvect^{\textrm{y}}+
\uvect^{\textrm{y}\top}\Vmat^{\textrm{yy}}\uvect^{\textrm{y}}>0
\hspace{5mm} \forall \hspace{2mm} \uvect^{\textrm{x}}\in\mathbb{R}^{L*},
\hspace{2mm} \forall \hspace{2mm} \uvect^{\textrm{y}}\in\mathbb{R}^{D*}.
$$
In particular, for $\uvect^{\textrm{x}}=-\Vmat^{\textrm{xx}-1}\uvect^{\textrm{y}}\Vmat^{\textrm{xy}}$ we obtain
$$
\uvect^{\textrm{y}\top}(\Vmat^{\textrm{yy}}-\Vmat^{\textrm{xy}\top}\Vmat^{\textrm{xx}-1}\Vmat^{\textrm{xy}})\uvect^{\textrm{y}}>0
\hspace{2mm}\Leftrightarrow\hspace{2mm} \uvect^{\textrm{y}\top}\Sigmamat\uvect^{\textrm{y}}>0
\hspace{5mm} \forall \hspace{2mm} \uvect^{\textrm{y}}\in\mathbb{R}^{D*}
$$
and hence $\Sigmamat\in\mathcal{S}^{D}_+$. $\blacksquare$
\begin{lemma}
\label{lemma:lemma2}
If
 $\Amat\in\mathbb{R}^{D\times L},
  \Gammamat\in\mathcal{S}^{L}_+,
  \Sigmamat\in\mathcal{S}^{D}_+$, then
  $$\Vmat=\left[
  \begin{array}{cc}
   \Gammamat & \Gammamat\Amat\tp\\
   \Amat\Gammamat & \Sigmamat+\Amat\Gammamat\Amat\tp
  \end{array}
 \right]\in\mathcal{S}^{L+D}_+.
$$
\end{lemma}
\paragraph{Proof.}
Since $\Gammamat\in\mathcal{S}^{L}_+$ there is a unique symmetric positive definite matrix $\Lambdamat\in\mathcal{S}^{L}_+$ such that $\Gammamat=\Lambdamat^2$. Using standard algebra, we obtain that for all non null $\uvect=[\uvect^{\textrm{x}};\uvect^{\textrm{y}}]\in\mathbb{R}^{L+D*}$,
$$
\uvect\tp\Vmat\uvect=||\Lambdamat\uvect^{\textrm{x}}+\Lambdamat\Amat\tp\uvect^{\textrm{y}}||^2+\uvect^{\textrm{y}\top}\Sigmamat\uvect^{\textrm{y}}
$$
where $||.||$ denotes the standard Euclidean distance. The first term of the sum is positive for all $[\uvect^{\textrm{x}};\uvect^{\textrm{y}}]\in\mathbb{R}^{L+D*}$ and the second term strictly positive for all $\uvect^{\textrm{y}}\in\mathbb{R}^{D*}$ because $\Sigmamat\in\mathcal{S}^{D}_+$ by hypothesis. Therefore, $\Vmat\in\mathcal{S}^{L+D}_+$. $\blacksquare$

Lemma~\ref{lemma:lemma1} and the correspondence formulae (\ref{eq:GLMtoJGMM}) prove
{\bf (i)}, Lemma~\ref{lemma:lemma2} and the correspondence formulae
(\ref{eq:JGMMtoGLM}) prove {\bf (ii)}, hence completing the
proof. $\blacksquare$
\section{The Marginal Hybrid GLLiM-EM}
\label{app:marginal} By marginalizing out the hidden variables
$\Wvect_{1:N}$, we obtain a different EM algorithm than the one
presented in section \ref{sec:algo}, with hidden variables
$\Zvect_{1:N}$ only. \textcolor{black}{For a clearer connection with
standard procedures we assume here, as already specified, that
$\cvect_k^{\textrm{w}} = \zerovect_{L_{\textrm{w}}}$ and
$\Gammamat_k^{\textrm{w}}= \Imat_{L_{\textrm{w}}}$.}
 The \textbf{E-W-step}
disappears while the \textbf{E-Z-step} and the following updating
of $\pi_k$, $\cvect_k^{\textrm{t}}$ and $\Gammamat_k^{\textrm{t}}$
in the \textbf{M-GMM-step} are exactly the same as in section
\ref{subsec:algo}. However, the marginalization of $\Wvect_{1:N}$
leads to a clearer separation between the regression parameters
$\Amat_k^{\textrm{t}}$ and $\bvect_k$ (\textbf{M-regression-step})
and the other parameters $\Amat_k^{\textrm{w}}$ and $\Sigmamat_k$
(\textbf{M-residual-step}). 
\textcolor{black}{
This can be seen
straightforwardly from equation (\ref{eq:model_ytwzbis}) which
shows that after marginalizing $\Wvect$, the model parameters
separate into a standard regression part $\Avect_k^{\textrm{t}}
\tvect_n + \bvect_k$ for which standard estimators do not involve
the noise variance and a PPCA-like part on the regression
residuals $\yvect_n- \tilde{\Avect}_k^{\textrm{t}} \tvect_n -
\tilde{\bvect}_k$, in which the non standard noise
covariance $\Sigmamat_k + \Amat_k^{\textrm{w}}
(\Amat_k^{\textrm{w}})\tp$ is typically dealt with by adding a latent
variable $\Wvect$. 
} 

 The algorithm
is therefore made of the {\bf E-Z-step} and {\bf M-GMM-step}
detailed in \ref{subsec:algo}, and the following M-steps:

\begin{description}
\item[\textbf{M-regression-step:}]
The $\Amat_k^{\textrm{t}}$ and $\bvect_k$ parameters are obtained using standard
weighted affine regression from $\{\tvect_n\}_{n=1}^N$ to
$\{\yvect_n\}_{n=1}^N$ with weights $\widetilde{r}_{nk}$, \textit{i.e.},
\begin{equation}
\widetilde{\Amat}^{\textrm{t}}_k=\widetilde{\Ymat}_k\widetilde{\Tmat}_k\tp(\widetilde{\Tmat}_k\widetilde{\Tmat}_k\tp)^{-1},\hspace{3mm}
   \widetilde{\bvect}_k = \sum_{n=1}^N\frac{\widetilde{r}_{kn}}{\widetilde{r}_{k}}(\yvect_n-\widetilde{\Amat}_k^t\tvect_{n})
\end{equation}
   with
$\widetilde{\Tmat}_k =
\left[\sqrt{\widetilde{r}_{1k}}(\tvect_1-\widetilde{\tvect}_k)
\dots \sqrt{\widetilde{r}_{Nk}}(\tvect_N-\widetilde{\tvect}_k)\right]\black{/\sqrt{\widetilde{r}_k}}$,
 $\widetilde{\tvect}_k =
 \sum_{n=1}^N\black{(\widetilde{r}_{kn}/\widetilde{r}_k)}\tvect_n\;.$

\item[\textbf{M-residual-step:}]
Optimal values for $\Amat^{\textrm{w}}_k$ and $\Sigmamat_k$ are
obtained by minimization of the  following criterion:
\begin{equation}
Q_k(\Sigmamat_k,\Amat_k^{\textrm{w}})= - \frac{1}{2}\biggl(\log|\Sigmamat_k+\Amat_k^{\textrm{w}}\Amat_k^{\textrm{w}\top}|
 +\sum_{n=1}^N\uvect_{kn}\tp(\Sigmamat_k+\Amat_k^{\textrm{w}}\Amat_k^{\textrm{w}\top})^{-1}\uvect_{kn}\biggr)
\end{equation}
\end{description}
where
$\uvect_{kn}=\sqrt{\widetilde{r}_{nk}/\widetilde{r}_k}(\yvect_n-\widetilde{\Amat}^{\textrm{t}}_k\tvect_n-\widetilde{\bvect}_k)$.
Vectors $\{\uvect_{kn}\}_{n=1}^N$ can be seen as the
\textit{residuals} of the  $k$-th local affine transformation. No
closed-form solution exists in the general case.
\textcolor{black}{A first option} is to make use of an inner loop
such as a gradient descent technique, or to consider $Q_k$ as the
new target observed-data likelihood and use an inner EM
corresponding to the general EM described in previous section with
$L_{\textrm{t}}=0$ and $K=1$. \textcolor{black}{Another option is
to use the Expectation Conditional Maximization (ECM) algorithm
\citep{MengRubin1993} proposed by \citep{ZhaoYu2008}. The ECM
algorithm replaces the M-step of the EM algorithm with a sequence
of conditional maximization (CM) steps. Such CM steps lead, in
the general case, to a conditional (to $\Sigmamat_k$) update of
$\Amat_k^{\textrm{w}}$ which is similar to PPCA
\citep{tipping1999probabilistic} with an isotropic noise variance
provided and equal to 1. It follows very convenient closed-form
expressions \citep{ZhaoYu2008} as is detailed below.
\citep{ZhaoYu2008} shows that such an ECM algorithm is
computationally more efficient than EM in the case of large sample
size relative to the data dimension and that the reverse may as
well be true in other situations. }

However, in the particular case $\Sigmamat_k=\sigma^2_k\Imat_D$,
\textcolor{black}{we can afford a standard EM as it connects to
PPCA. Indeed,} one may notice that $Q_k$ has then exactly the same
form as the observed-data log-likelihood in PPCA, with parameters
$(\sigma^2_k,\Amat_k^{\textrm{w}})$ and observations
$\{\uvect_{kn}\}_{n=1}^N$. Denoting with $\Cmat_k= \sum_{n=1}^N
\uvect_{kn} \uvect_{kn}^T\black{/N}$ the $D\times D$ sample \textit{residual
covariance matrix} and with $\lambda_{1k}>\dots>\lambda_{Dk}$ its
eigenvalues in decreasing order, we can therefore use the key
result of \citep{tipping1999probabilistic} to see that a global
maximum of $Q_k$ is obtained for
\begin{align}
 \widetilde{\Amat}_k^{\textrm{w}} &=\Umat_k(\Lambdamat_k-\sigma^2_k\Imat_{L_{\textrm{w}}})^{1/2},\\
  \widetilde{\sigma}^2_k &=\frac{\sum_{d=L_{\textrm{w}}+1}^D\lambda_{dk}}{D-L_{\textrm{w}}}
\end{align}
where $\Umat_k$ denotes the $D\times L_{\textrm{w}}$ matrix whose
column vectors are the first eigenvectors of $\Cmat_k$ and
$\Lambdamat_k$ is a $L_{\textrm{w}}\times L_{\textrm{w}}$ diagonal
matrix containing the corresponding first eigenvalues. 

The hybrid nature of hGLLiM (at the crossroads of regression and dimensionality reduction) is striking in this variant, as it alternates between a
mixture-of-Gaussians step, a local-linear-regression step and a
local-linear-dimensionality-reduction step on residuals. This
variant is also much easier to initialize as a set of initial
posterior values $\{r_{nk}^{(0)}\}_{n=1,k=1}^{N,K}$ can be
obtained using the $K$-means algorithm or the standard GMM-EM
algorithm on $\tvect_{1:N}$ or on the joint data $[\yvect;\tvect]_{1:N}$ as done in
\citep{QiaoMinematsu09} before proceeding to the M-step. On the
other hand, due to the time-consuming eigenvalue decomposition needed at each
iteration, the marginal hGLLiM-EM turns out to be slower that the general hGLLiM-EM algorithm
described in section \ref{sec:algo}. We
thus use the marginal algorithm as an
initialization procedure for the general hGLLiM-EM algorithm.

\bibliographystyle{spbasic}


\end{document}